%% file: main.tex
\documentclass{article}

\usepackage[main, final]{neurips_2026}

\usepackage[utf8]{inputenc} % allow utf-8 input
\usepackage[T1]{fontenc}    % use 8-bit T1 fonts
\usepackage{hyperref}       % hyperlinks
\usepackage{url}            % simple URL typesetting
\usepackage{booktabs}       % professional-quality tables
\usepackage{amsfonts}       % blackboard math symbols
\usepackage{nicefrac}       % compact symbols for 1/2, etc.
\usepackage{microtype}      % microtypography
\usepackage{xcolor}         % colors
\usepackage[table]{xcolor}
\usepackage{wrapfig}
\usepackage{xcolor}
\usepackage{graphicx}
\usepackage{caption}
\usepackage{multirow}
\usepackage{subcaption}
\usepackage{enumitem}
\usepackage{float}
\usepackage{caption}
\usepackage[export]{adjustbox}

\usepackage[most]{tcolorbox}
\usepackage{listings}
\usepackage{fvextra}
\tcbuselibrary{listings,breakable}

\newtcolorbox{promptbox}{
    breakable,
    enhanced jigsaw,
    colback=gray!3,
    colframe=black,
    boxrule=0.4pt,
    arc=1mm,
    left=1mm,
    right=1mm,
    top=1mm,
    bottom=1mm,
    width=\linewidth
}
\definecolor{beige}{RGB}{234,240,206}
\definecolor{silver}{RGB}{192,197,193}
\definecolor{grey}{RGB}{125,132,145}
\definecolor{lightpurple}{RGB}{87, 75, 96}
\definecolor{batchblue}{RGB}{40,100,160}
\definecolor{darkpurple}{RGB}{63, 51, 77}
\definecolor{darkred}{rgb}{0.8, 0.25, 0.33}

\definecolor{lightgray}{RGB}{245,245,245}
\definecolor{midgray}{RGB}{230,230,230}

\title{AgentSSL: Can MLE Agents Leverage Unlabeled Data?}
\author{
Ethan Lin$^{1}$\thanks{Equal contribution.} \quad
Akanksha Sarkar$^{1}$\footnotemark[1] \quad
Ziang Liu$^{1}$ \quad
Kristin Branson$^{2}$ \quad
Jennifer J. Sun$^{1}$
\\[1ex]
$^{1}$Cornell University \qquad
$^{2}$Howard Hughes Medical Institute
}

\begin{document}
\maketitle

\begin{abstract}

Applying machine learning systems to new domains introduces bottlenecks in both labeled data acquisition and manual pipeline engineering. While recent MLE agents show promise in automating pipeline design, they have been studied almost exclusively in fully supervised settings, leaving them untested in realistic regimes where labeled data is scarce but unlabeled data is abundant. In this work, we introduce Agentic Semi-Supervised Learning (\textit{AgentSSL}), framing SSL as a program search problem and studying how agents navigate this space under limited supervision. We find that even under weak feedback and matched compute, agents can synthesize code that combines and adapts established SSL techniques, exceeding the strongest human-engineered baselines in 8 of 12 settings by up to +21.8\%. However, our results identify program selection as a critical bottleneck in AgentSSL, where performance drops by as much as 47\% when selecting based on the agent's feedback metric. To systematically study this bottleneck, we introduce \textit{ProgramSelectBench}, a large-scale benchmark with a combined total of 16,977 agent-generated and human-engineered SSL programs, and evaluate both unsupervised and active program selection methods. We further evaluate AgentSSL on real-world ecology detection tasks, highlighting the opportunities and challenges of transferring agent-discovered strategies beyond controlled benchmarks.

\footnotetext[1]{Project website: \url{https://codiscovery.ai/AgentSSL}}
\footnotetext[2]{Selection Benchmark: \url{https://huggingface.co/datasets/ethan-y-lin/ProgramSelectBench}}

\end{abstract}

\input{sec/1_intro}
\input{sec/2_related_works}
\input{sec/3_problem_setup}
\input{sec/4_experiments_details}

\input{sec/5_results}

\input{sec/6_analysis}
\input{sec/7_ablations}
\input{sec/8_ecology_application}
\input{sec/9_conclusion}

\clearpage
\section*{Acknowledgements}
This material is based on work supported by the AI Research Institutes program supported by the NSF and Intel Corporation under NSF award DMR-2433348.
This research is supported in part by the Cornell Bowers Undergraduate Research Experience (BURE) program and the National Science Foundation (IIS-2505098). Research supported by the NVIDIA Academic Grant Program using 4 × RTX PRO 6000 Blackwell GPUs.

\bibliographystyle{plain}  % or unsrt, alpha, ieeetr, etc.
\bibliography{references}
\clearpage
%%%%%%%%%%%%%%%%%%%%%%%%%%%%%%%%%%%%%%%%%%%%%%%%%%%%%%%%%%%%
\appendix
\input{sec/appendix}
\clearpage 
\end{document}

%% file: sec/1_intro.tex
\vspace{-1.5em}
\section{Introduction}
\begin{wrapfigure}{r}{0.50\linewidth}
    \vspace{-3.9em}
    \centering
    \includegraphics[width=\linewidth]{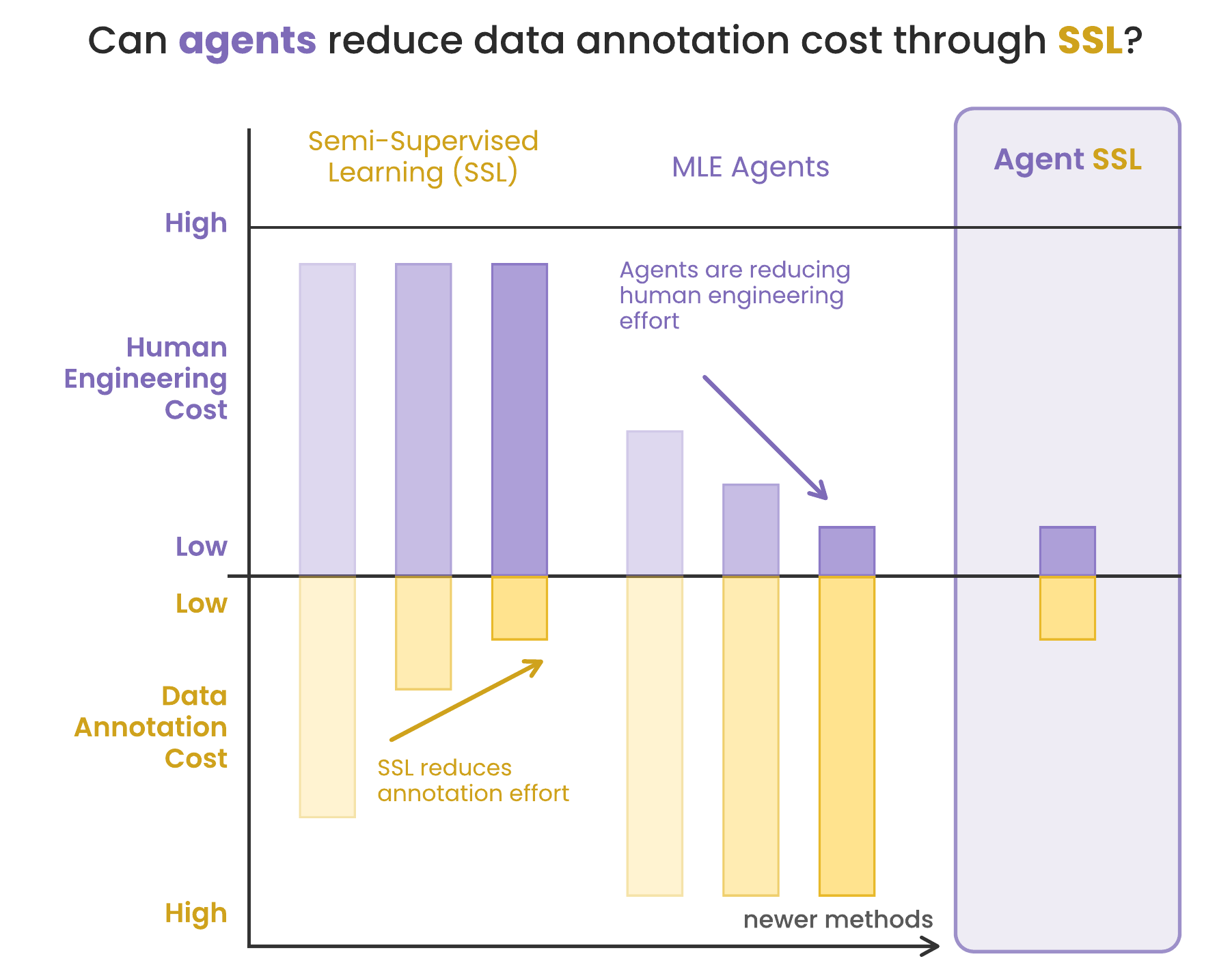}
    \caption{\small MLE Agent research is focused on reducing engineering cost, but relies on human labeled data. SSL research seeks to reduce annotation effort but requires human engineering to deploy. We study whether \textbf{AgentSSL} can reduce both costs.}
    \vspace{-2.5em}
    \label{fig:intro}
\end{wrapfigure}

\textit{“What we observe is not nature itself, but nature exposed to our method of questioning.” \\
— Werner Heisenberg}

Modern machine learning depends on two costly inputs: (1) \textbf{labeled data} to provide supervision, and (2) \textbf{human engineering} to design the inductive biases and pipelines (e.g. architectures, augmentations, and training strategies) required to extract signal from that data \cite{hestness2017deep, kaplan2020scaling}. 

Recent agent-based approaches target engineering cost by reframing ML pipeline development as program search, iteratively generating and refining executable code based on task feedback. The capabilities of these agents have improved significantly. For example, on MLE-Bench~\cite{chan2024mle}, a benchmark of 75 Kaggle competitions, the top agent performance has risen from a mean percentile rank of 45.2\% \cite{toledo2025ai} to 83.1\% \cite{hambardzumyan2026aira_2} in just a single year, driven by a rapidly expanding ecosystem of agents~\cite{jiang2025aideaidrivenexplorationspace,liu2025mlmasteraiforaiintegrationexploration,nam2025mlestarmachinelearningengineering,zhang2026aibuildaiaiagentautomatically,hambardzumyan2026aira_2}. Despite this success, there remains a fundamental limitation in our \textit{method of questioning}: existing agentic benchmarks overwhelmingly assume fully supervised settings. MLE-Bench primarily contains tasks with thousands to millions of training samples. As a result, agents are optimized to search over programs under the assumption that labeled data is abundant.

This assumption rarely holds in practice. In many real-world domains (including scientific discovery, medical diagnostics, and environmental monitoring), labeled data is scarce and expensive while unlabeled data is abundant. This raises a fundamental question: \textit{can agent-based systems remain effective in data-scarce regimes by leveraging unlabeled data?}

This places the problem naturally within semi-supervised learning (SSL), where a small labeled set is paired with a much larger unlabeled corpus. Despite strong results on classical benchmarks such as CIFAR-10 and CIFAR-100 \cite{krizhevsky2009learning}, SSL remains difficult to deploy in practice since these methods are highly sensitive to how well the data distribution matches the algorithm's underlying assumptions (e.g. cluster structure, consistency under augmentation, etc.) \cite{zhu2005semi, Yang_2023}. As a result, the methods often require significant dataset-specific tuning. This brittleness suggests that deploying SSL can be viewed as a search problem over a large design space spanning graph-based label propagation~\cite{zhu2005semi}, pseudo-labeling~\cite{pseudolabeling}, consistency regularization~\cite{tarvainen2017mean}, and many more \cite{zhu2005semi, Yang_2023}.

In this work, we introduce \textbf{Agentic Semi-Supervised Learning (AgentSSL)}, formulating the domain-specific application of SSL as a program search problem. We decompose this problem into two core challenges: \textbf{(1) program discovery}, where agents must discover effective SSL programs under limited feedback, and \textbf{(2) program selection}, where a single program must be identified from the discovered candidates without a large labeled validation set.

To study program discovery, we investigate two unsupervised feedback approaches for guiding agentic search. We find that, even when unsupervised feedback is weakly or even negatively correlated with final accuracy, agents can still discover SSL programs that are competitive with, and in 8 of 12 settings \textbf{outperform}, recent SSL baselines. Through extensive empirical and qualitative analysis, we find that agents explore and combine diverse ideas spanning decades of SSL research, with the strongest agent-generated solutions effectively leveraging both unlabeled data and domain-specific structure.

Despite discovering strong SSL programs, agents face a significant challenge in reliably selecting the best programs. This reflects a longstanding challenge in practical SSL: selecting methods and hyperparameters without access to a reliable labeled validation set \cite{oliver2019realisticevaluationdeepsemisupervised,zhang2025revisiting}. We evaluate two promising approaches. First, we find that unsupervised metrics enable effective selection when underlying assumptions hold. Second, when unsupervised metrics fail, actively allocating labels to program selection can outperform allocating them to training. To facilitate further research, we introduce \textbf{ProgramSelectBench} (Appendix~\ref{appendix:selection}), a large-scale program selection benchmark comprising 16,977 precomputed programs representing over 10,000 GPU hours of compute.

Finally, we evaluate AgentSSL on animal detection tasks from real-world ecology pipelines, showing both its potential to reduce annotation cost and the instability of deploying agent-discovered SSL strategies outside controlled classification benchmarks. To summarize, our contributions are: 
\begin{itemize}[leftmargin=1.2em,topsep=2pt,itemsep=2pt,parsep=0pt,partopsep=0pt]
    \item We introduce \textbf{Agentic Semi-Supervised Learning} (AgentSSL), formulating SSL as agentic search over the space of SSL programs. 

    \item We show that MLE agents can autonomously discover diverse, domain-adapted SSL programs that effectively leverage unlabeled data, achieving performance competitive with established SSL baselines, outperforming them in 8 of 12 settings.

    \item We identify reliable program selection as an open problem for AgentSSL, and release \textbf{ProgramSelectBench}, a large-scale program selection benchmark with 16,977 programs of human baseline and agent solutions. 
    
    \item We evaluate AgentSSL on real-world ecology detection tasks, demonstrating the potential and challenges for agents to reduce annotation-cost in practice.
\end{itemize}

%% file: sec/2_related_works.tex
\vspace{-0.5em}
\section{Related Work}
\vspace{-0.5em}
\subsection{Automating Machine Learning Engineering (MLE)}

Automating MLE has long been a goal of the ML community. Traditional AutoML methods, including hyperparameter optimization \cite{akiba2019optunanextgenerationhyperparameteroptimization}, pipeline search \cite{TPOT2019}, and neural architecture search (NAS) \cite{white2023neuralarchitecturesearchinsights,He_2021}, typically formulate automation as optimization over predefined search spaces, such as model architectures, training procedures, and hyperparameters. While effective in many settings, these methods often require substantial human specification of the search space and optimization objective.

Recent LLM-based MLE agents offer a more flexible alternative by searching over executable programs and iteratively proposing, executing, debugging, and refining code using feedback from the environment \cite{novikov2025alphaevolvecodingagentscientific,jiang2025aideaidrivenexplorationspace,liu2025mlmasteraiforaiintegrationexploration,nam2025mlestarmachinelearningengineering,zhang2026aibuildaiaiagentautomatically,hambardzumyan2026aira_2}. Because these agents combine many interacting design choices, controlled evaluation frameworks are important for isolating the effects of search policies, feedback signals, and other components \cite{toledo2025ai,yang2025rdagentllmagentframeworkautonomous}. We use Aira-Dojo \cite{toledo2025ai} as such a test harness, providing a consistent environment for studying these design choices in our SSL setting.

Existing MLE agents are also typically evaluated on supervised-learning benchmarks with substantial labeled training and validation data. MLE-Bench \cite{chan2024mle} evaluates agents across 75 Kaggle competitions, while AIRS-Bench \cite{lupidi2026airsbench} considers challenging tasks drawn from ML research papers. In contrast, we evaluate MLE agents in a data-efficient setting where labeled data is limited, making program discovery and program selection under weak feedback itself a central challenge.

\subsection{Semi-Supervised Learning (SSL)}

SSL studies how to leverage unlabeled data together with limited labeled data. We briefly review its major paradigms since our agent builds on both traditional and modern SSL methods.

\textbf{Classical SSL methods.}
Classical methods impose assumptions on how labeled and unlabeled examples should relate~\cite{zhu2005semisupervised}. For example, graph-based methods encourage connected examples on a similarity graph to share labels \cite{song2021graphbasedsemisupervisedlearningcomprehensive}; low-density separation methods, such as transductive SVMs, encourage decision boundaries to avoid dense regions \cite{pmlr-vR5-chapelle05b}.

\textbf{Deep SSL methods.}
Deep SSL implements similar assumptions through neural training objectives \cite{Yang_2023}. Pseudo-labeling and self-training use model predictions on unlabeled data as targets. Consistency regularization enforces invariance to augmentations. Early methods like $\Pi$-Model \cite{laine2017temporalensemblingsemisupervisedlearning} and Mean Teacher \cite{tarvainen2018meanteachersbetterrole} used consistency between different trained networks as an objective. FixMatch \cite{sohn2020fixmatchsimplifyingsemisupervisedlearning} combines several SSL approaches, including weakly augmented pseudo-labeling, confidence thresholding, and strongly augmented consistency training. Subsequent methods such as SoftMatch \cite{chen2023softmatchaddressingquantityqualitytradeoff} refine this through adaptive thresholds, curriculum selection, or confidence weighting. FineSSL and V-PET are recent SSL methods tailored for foundation models, either through addressing biases in foundation-model adaptation \cite{gan2024erasingbiasfinetuningfoundation} or ensembling complementary pseudolabels from multiple PEFT methods and foundation-model backbones \cite{zhang2025revisiting}.

\subsection{Unsupervised and Active Model Selection}

Model selection without a large labeled validation set remains an open challenge in data-efficient settings. \textit{Unsupervised model selection} replaces labeled validation performance with statistics computed from model predictions or learned representations. Many such metrics were originally developed in other contexts, including classical clustering and statistics \cite{fowlkes1983method,vinh2009information,Halkidi2002ClusterVM,Caliski1974ADM}, representation learning \cite{garrido2023rankmeassessingdownstreamperformance}, and domain adaptation or data-efficient learning \cite{cui2020discriminabilitydiversitybatchnuclearnorm,saito2021tune}. Recent work has repurposed and evaluated these signals as unsupervised validation criteria in unsupervised domain adaptation \cite{ericsson2023betterpracticesdomainadaptation} and SSL \cite{zhang2025revisiting}, but their reliability depends on assumptions that may not hold consistently across datasets or models \cite{kay2025consensusdrivenactivemodelselection, ericsson2023betterpracticesdomainadaptation}. An orthogonal direction is \emph{active model selection}, which uses a limited annotation budget to identify the best model. Existing approaches include uncertainty- or variance-driven methods that prioritize labels most useful for reducing uncertainty in model performance estimates \cite{kossen2021activetestingsampleefficientmodel,matsuura2023active}, as well as probabilistic approaches that model uncertainty over candidate model quality and use this posterior to guide label acquisition \cite{kay2025consensusdrivenactivemodelselection, okanovic2024modelswrongusefulmodel}. Both directions provide promising avenues for improving model selection in AgentSSL when a large labeled validation set is unavailable.

\textbf{Benchmarking model selection.} Existing evaluations of model selection under limited supervision remains limited, both in size and diversity of candidate models. In particular, active model selection methods are typically evaluated over a finite set derived from following the same fine-tuning procedure on different pre-trained models and hyperparameters \cite{okanovic2024modelswrongusefulmodel, kay2025consensusdrivenactivemodelselection}. This leaves open how well existing selection methods scale to settings where candidate solutions may differ substantially in architectures, training procedures, and hyperparameters. 

%% file: sec/3_problem_setup.tex
\vspace{-0.5em}
\section{Problem Setup}
\vspace{-0.5em}

\begin{figure}[!t]
    \centering
    \includegraphics[
        width=\linewidth]{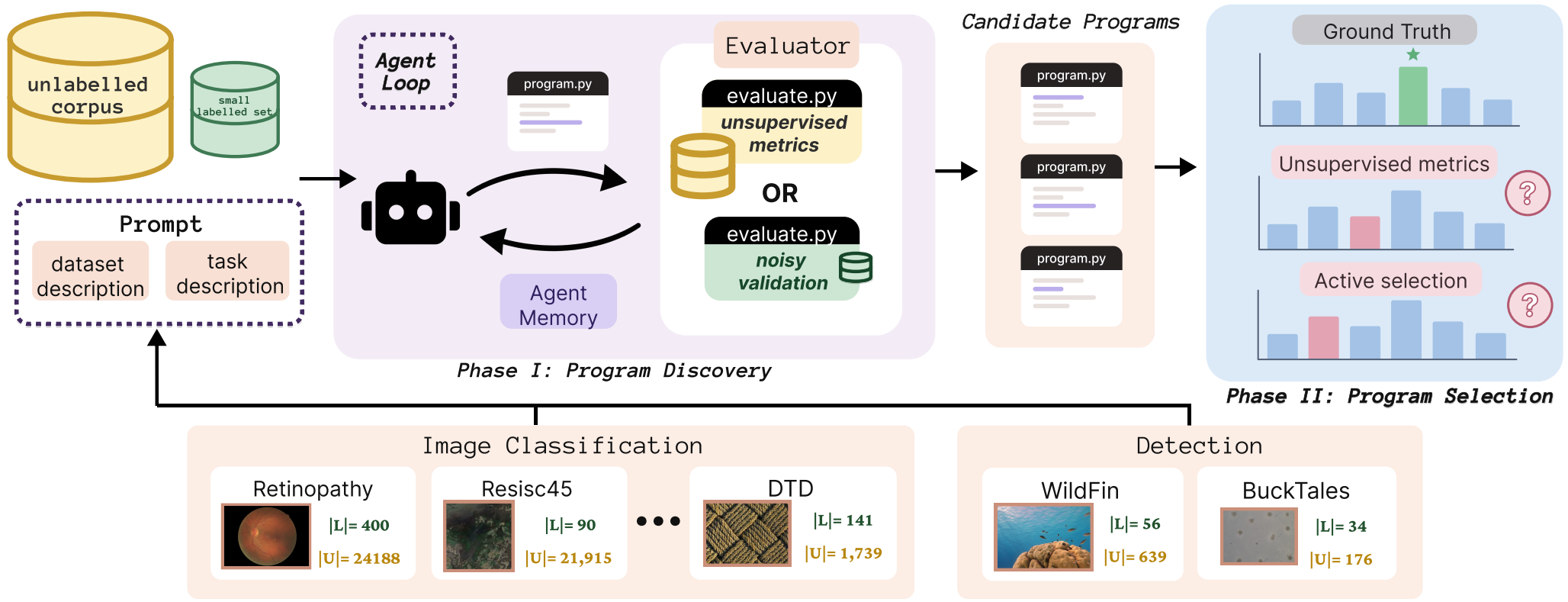}
    \caption{\small Overview of AgentSSL. Given a small labeled set and a larger unlabeled corpus, an LLM-based agent generates SSL programs that are evaluated using either unsupervised SSL metrics or noisy validation feedback, with the resulting fitness score guiding iterative refinement. Following search, we must select the best program either using unsupervised metrics or active model selection. We evaluate AgentSSL on both image classification SSL benchmarks and real-world ecological detection tasks.}
    \label{fig:agent_main_figure}
\end{figure}

\subsection{SSL Problem Formulation}
In the SSL setting, each task provides a labeled dataset $L = \{(x_i^\ell, y_i^\ell)\}_{i=1}^{|L|}$
and an unlabeled dataset $U = \{x_i^u\}_{i=1}^{|U|},$ where typically $|L| \ll |U|$ with the goal of constructing a prediction function $ h : \mathcal{X} \rightarrow Y$ that achieves high performance on a held-out test distribution.

Rather than restricting the solution space to a particular parameterized model trained with a fixed SSL objective, we define a solution more generally as a program
$ \mathcal{A}: (L, U) \mapsto h,$
which takes the labeled and unlabeled data as input and returns a predictor. This formulation includes conventional deep SSL methods that learn parameters $\theta$ of a model $f_\theta$ using both $L$ and $U$, while also allowing broader strategies such as classical transductive or graph-based SSL methods.

\subsection{AgentSSL Problem Formulation}

We formulate AgentSSL as an agentic search problem over the space of SSL programs, comprising two core challenges: \textbf{(1) program discovery}, where an agent iteratively generates and refines candidate programs under limited feedback, and \textbf{(2) program selection}, where a single program must be identified from the discovered candidates for deployment under label scarcity.

\textbf{Program Discovery}. We define the solution space as a set of programs $\mathcal{S} = \{\mathcal{A} : (L,U) \mapsto h\}$, where each program specifies a complete learning procedure for constructing a predictor from the available data. Following prior work \cite{toledo2025ai}, we model agentic search as growth of a directed acyclic graph over programs: greedy and MCTS expand single nodes and induce trees, while evolutionary crossover combines two parents into one child. At iteration $t$, the agent maintains a directed graph $ G_t = (V_t, E_t),$ where each node $v \in V_t \subseteq \mathcal{S}$ represents a candidate SSL program and each directed edge $(v_i, v_j) \in E_t$ represents a transformation from one program to another. At each iteration, the agent selects one or more existing programs, applies an operator to produce a new candidate program, evaluates it using a feedback signal, and adds the resulting node to the search graph. Concretely, an AgentSSL search procedure is specified by the tuple
$\left(F, \pi_{\mathrm{sel}}, \mathcal{O}, \pi_{\mathrm{op}}, M\right),$ where $F$ is a fitness function, $\pi_{\mathrm{sel}}$ is a node selection policy, $\mathcal{O}$ is a set of program transformation operators, $\pi_{\mathrm{op}}$ is an operator selection policy, and $M$ is the LLM. We use a fixed operator set with Draft, Debug, Improve, and Analyze operators (Appendix~\ref{appendix:operators}). 

\textbf{Program Selection}.
Following program discovery, we obtain a pool of candidate
programs $\mathcal{P} \subseteq \mathcal{S}$. The goal of
program selection is to identify the best-performing program
$\mathcal{A}^* \in \mathcal{P}$, defined as
$\mathcal{A}^* = \arg\max_{\mathcal{A} \in \mathcal{P}} \mathcal{Q}(\mathcal{A}),$
where $\mathcal{Q}(\mathcal{A})$ denotes the performance of the predictor produced by program $\mathcal{A}$ on the test distribution. However, true performance is unknown, and the limited labeled data available in SSL makes it difficult to reliably identify $\mathcal{A}^*$. We therefore study program selection under different information constraints. \textbf{Unsupervised selection} chooses programs using label-free metrics \cite{zhang2025revisiting} computed on an unlabelled validation set. \textbf{Active model selection} chooses programs by actively querying a limited number of test labels.

\textbf{Evaluation}. We evaluate these two challenges separately. To assess program discovery, we measure \textit{oracle performance}, defined as the test performance of the best program in the discovered pool, $\mathcal{Q}(\mathcal{A}^*)$. This measures the quality of the best discovered solution, independent of our ability to select it. To assess program selection, we measure \textit{selection regret}, defined as $R_{\mathrm{sel}} = \mathcal{Q}(\mathcal{A}^*) - \mathcal{Q}({\mathcal{A}}_s),$ where ${\mathcal{A}}_s$ is the program chosen by a selection method. Lower regret indicates that the selection method more reliably identifies high performing programs from the discovered pool.

\subsection{Agent Design Space}

The AgentSSL formulation exposes several design choices that can affect how agents explore the space of SSL programs. We focus on three components that are especially relevant to our study: the fitness function, the search strategy, and the language model.

\textbf{Fitness Functions ($F$).} The fitness function defines the feedback signal received by the agent during search. This is especially important in SSL, where labeled validation data is scarce. We consider two families of feedback signals: (i) \textbf{noisy validation}, where a small labeled subset is held out for evaluation, and (ii) \textbf{unsupervised metrics}, \cite{zhu2005semi,chapelle2009semi,zhang2025revisiting} where programs are scored using label-free statistics computed from predictions or representations on unlabeled data. Noisy validation provides direct supervised feedback but can have high variance in low-label regimes. Unsupervised metrics instead test whether agents can search effectively using only structure in unlabeled data. The construction of the noisy validation signal introduces a tradeoff between robustness and evaluation cost: stronger estimates such as cross-validation reduce variance but substantially increase program evaluation time. In our experiments, we tune this configuration separately for each dataset, though we view automated adaptation of this tradeoff as an interesting direction for future work. More details about the metric calculation in Appendix~\ref{appendix:feedback_metrics}.

\textbf{Search Strategies ($\pi_{\mathrm{sel}}$).}
The search policy $\pi_{\mathrm{sel}}$ determines which nodes $v \in V_t$ in the search graph $G_t = (V_t, E_t)$ are expanded at each iteration. Different strategies induce different exploration--exploitation tradeoffs: greedy selects the highest-score programs under $F$, evolutionary search mutates a population of candidate programs, and MCTS balances exploration and exploitation through tree search. We study how $\pi_{\mathrm{sel}}$ affects performance and diversity of discovered SSL strategies.

\textbf{Language Model Backbone ($M$).}
The language model $M$ instantiates the program transformation operators $\mathcal{O}$ by generating modifications to existing programs. Given a node $v \in V_t$ and operator $o \in \mathcal{O}$, the model proposes a transformed program $v'$. Differences in coding ability and prior SSL knowledge can therefore affect both the quality and diversity of discovered solutions.

\textbf{AgentSL} To establish a simple agentic baseline, we perform experiments using agentic supervised learning (AgentSL) by removing the unlabeled data as input to the agent.  

%% file: sec/4_experiments_details.tex
\vspace{-0.5em}
\section{Experiment Design}
\vspace{-0.5em}

\textbf{Datasets.} We conduct our experiments using the image classification SSL setting from recent work on semi-supervised learning with foundation models \cite{zhang2025revisiting}. Following the VTAB protocol \cite{zhai2019large}, it includes six datasets across three categories: \textit{Natural} (DTD \cite{cimpoi2014describing}, SUN397 ~\cite{xiao2010sun}), \textit{Specialized} (RESISC45 \cite{cheng2017remote}, Retinopathy (RETINO) \cite{kaggleDiabeticRetinopathy}), and \textit{Structured} (CLEVR-Count (CLEVR) \cite{johnson2017clevr}, KITTI ~\cite{geiger2013vision}), spanning texture recognition, scene understanding, remote sensing, medical imaging, synthetic reasoning, and autonomous driving. For each dataset, we evaluate two class-balanced settings, corresponding to different numbers of labeled examples per class. For robustness, since few-shot results can be sensitive to which examples are labeled, we generate three random splits for each dataset and shot setting and evaluate all agent and baseline methods on each split.

\textbf{Baselines.}
We compare agents against six baselines: Supervised, FixMatch, SoftMatch, FineSSL, PET, and V-PET. Supervised baseline measures performance without unlabeled data; FixMatch and SoftMatch represent standard deep SSL methods; FineSSL, PET, and V-PET represents SSL strategies tailored for foundation models. For all baselines, we follow the original protocol from prior work \cite{zhang2025revisiting}, focusing on the subset of their experiments that use CLIP ViT-B-16 \cite{radford2021learningtransferablevisualmodels} and DINOv2-B \cite{oquab2024dinov2learningrobustvisual} and fine-tune with LoRA \cite{hu2021loralowrankadaptationlarge}. See Appendix~\ref{appendix:baseline_details} for more details. 

\textbf{Agent Implementation.} 
We build on AIRA-Dojo \cite{toledo2025ai} as an agentic program-search framework and introduce a new SSL task interface within it. Agents receive a task description together with the labeled set $L$ and unlabeled set $U$, and propose executable programs that map these data inputs to a predictor $h$ (Figure~\ref{fig:agent_main_figure}). Candidate programs are executed in an isolated Apptainer environment using a user-defined fitness evaluation program, and the resulting execution logs, errors, and fitness scores are returned to the agent as feedback for subsequent search steps.

\textbf{Agent Details.} 
To make agent-discovered programs directly comparable to the baselines, we constrain the agent search space in our image classification experiments. Agents thus only use CLIP ViT-B-16 and DINOv2-B backbones and are provided access to the same parameter-efficient fine-tuning library used in the baseline evaluations. We initialize each search with a simple supervised warm-start program to focus the agent on exploring SSL strategies rather than rediscovering the experimental infrastructure. We evaluate agents using two different feedback metrics, \textbf{U}nsupervised (\textbf{U}) and \textbf{N}oisy-validation (\textbf{N}), and use a greedy search policy. See Appendix~\ref{appendix:agent_details} for more details.

\textbf{Search Budget.} We allocate agent search budgets based on the computational cost of the corresponding baseline SSL methods for each dataset. Since candidate evaluation time varies substantially with dataset size, this scaling gives agents more wall-clock time on larger datasets. For the smaller datasets, DTD, RESISC45, and KITTI, we run each agent for 8 hours on an NVIDIA RTX 6000 Ada Generation GPU. For the remaining larger datasets, we run each agent for 15 hours on either an NVIDIA H100 GPU or an NVIDIA RTX 6000 Blackwell Max-Q GPU. For a comparison between baseline and agent compute costs, see Appendix~\ref{appendix:baseline_compute}.

\textbf{Evaluation Protocol.}
For each dataset and shot setting, we construct three random splits, corresponding to different labeled and unlabeled sets $(L,U)$. We evaluate every baseline method and hyperparameter configuration on all three splits, and run one independent agent trajectory on each split. We define the "best" program as the program with the best average performance across all data splits. Thus, every agent-generated program is executed and evaluated on the other two splits as well.

%% file: sec/5_results.tex
\section{Results}

\input{tables/1_main_table}

Across datasets and shot settings, AgentSSL discovers programs that are competitive with strong SSL baselines when programs are selected using oracle test performance (Table~\ref{tab:main_results}). This result shows that the agent search process is capable of discovering high-performing SSL programs even when feedback is weakly or even negatively correlated with test accuracy (Figure~\ref{fig:unsup_corr}).

\input{tables/2_unsup_regret}

The gap between oracle and practical selection, however, highlights a central challenge for AgentSSL and SSL more broadly: \textit{identifying the best discovered program when validation feedback is weak}. Table~\ref{tab:unsup_regret} reports practical selection regret when using the feedback metric for selection. When the metric is well correlated with test performance, much of the oracle performance is retained; when this alignment is weak, selection performance can degrade substantially. These results suggest that practical program selection is a major bottleneck for AgentSSL. We provide a comprehensive evaluation of all unsupervised selection metrics on ProgramSelectBench in Appendix~\ref{appendix:selection}.

A limitation of unsupervised selection is that its metrics necessarily encode assumptions about what constitutes a good model. For example, several metrics assess properties of learned representations. However, the agent may discover programs that reuse the same frozen representation while applying different downstream SSL procedures. In such cases, representation-based metrics cannot distinguish between programs despite potentially large differences in predictive performance.

\input{tables/3_active_selection_shotbridge}

When unsupervised metrics fail, active model selection offers a promising alternative by using labels for selection. Under a fixed budget, we compare training on all labels with unsupervised selection against splitting labels evenly between training and active selection (Table~\ref{tab:active_selection_shotbridge}). This allocation yields large gains where unsupervised metrics fail, though it costs accuracy where they already work. Active selection strategies carry their own tradeoffs. IID, ActiveTesting, and VMA perform well on average but have much higher variance across runs than probabilistic methods such as CODA (deterministic) and ModelSelector (Table~\ref{appendix:active-selection-std}). Conversely, CODA and ModelSelector can converge with fewer labels but can be unstable as labels accumulate (Figure~\ref{appendix:regret-vs-iteration}). We provide further evaluation and analysis of active model selection on ProgramSelectBench in Appendix~\ref{appendix:selection}.

%% file: tables/1_main_table.tex
\definecolor{purple}{HTML}{846DA3}
\definecolor{yellow}{HTML}{A18600}
\definecolor{midgray}{gray}{0.88}
\definecolor{plus}{HTML}{0B7A3B}
\definecolor{minus}{HTML}{B22222}

\newcommand{\human}[1]{\textbf{\textcolor{yellow}{#1}}}
\newcommand{\agent}[1]{\textbf{\textcolor{purple}{#1}}}
\newcommand{\plus}[1]{\textbf{\textcolor{plus}{#1}}}
\newcommand{\minus}[1]{\textbf{\textcolor{minus}{#1}}}

\begin{table*}[t]
\centering
\footnotesize
\setlength{\tabcolsep}{2.6pt}
\renewcommand{\arraystretch}{1.05}
\resizebox{0.98\textwidth}{!}{%
\begin{tabular}{@{}l*{12}{c}@{}}
\toprule
 & \multicolumn{2}{c}{\textbf{DTD}} & \multicolumn{2}{c}{\textbf{SUN397}} & \multicolumn{2}{c}{\textbf{RESISC45}} & \multicolumn{2}{c}{\textbf{RETINO}} & \multicolumn{2}{c}{\textbf{CLEVR}} & \multicolumn{2}{c}{\textbf{KITTI}} \\
\cmidrule(lr){2-3}\cmidrule(lr){4-5}\cmidrule(lr){6-7}\cmidrule(lr){8-9}\cmidrule(lr){10-11}\cmidrule(lr){12-13}
\textbf{Method} & 3 & 6 & 3 & 6 & 1 & 2 & 40 & 80 & 10 & 20 & 5 & 10 \\
\midrule
Supervised & 61.47 & 70.23 & 63.35 & 70.54 & 55.71 & 70.98 & \human{50.66} & 51.48 & 39.96 & 46.62 & \human{41.54} & \human{45.66} \\
FixMatch & 60.71 & 70.12 & 63.10 & 70.90 & 62.13 & 81.60 & 34.93 & 39.70 & 38.93 & 56.32 & 39.05 & 42.76 \\
SoftMatch & 62.59 & 69.80 & \human{69.68} & \human{73.29} & 73.17 & 86.04 & 44.14 & \human{54.37} & \human{46.48} & \human{66.83} & 40.04 & 41.30 \\
FineSSL & 65.83 & \human{73.00} & 68.41 & 72.33 & \human{76.13} & \human{86.93} & 27.47 & 27.43 & 34.21 & 47.54 & 40.74 & 37.22 \\
PET & 65.43 & 71.97 & 63.94 & 68.19 & 60.48 & 78.31 & 30.45 & 32.77 & 33.34 & 34.31 & 39.01 & 42.80 \\
V-PET & \human{66.03} & 72.87 & 66.87 & 71.92 & 60.94 & 76.99 & 24.12 & 29.13 & 38.07 & 35.17 & 39.29 & 41.02 \\
\rowcolor{lightgray}
$\mathrm{AgentSSL}^{\textbf{N}}$ & 65.50 & 72.38 & \agent{69.14} & 70.90 & -- & 79.82 & \agent{72.47} & \agent{73.47} & \agent{51.57} & \agent{52.92} & 48.52 & 50.21 \\
\rowcolor{lightgray}
$\mathrm{AgentSSL}^{\textbf{U}}$ & \agent{68.94} & \agent{73.76} & 68.46 & \agent{72.36} & \agent{79.80} & \agent{80.98} & 50.73 & 49.65 & 43.43 & 52.81 & \agent{49.74} & \agent{53.45} \\
\rowcolor{midgray}
\agent{$\text{A}_{\text{best}}$}$\,-\,$\human{$\text{H}_{\text{best}}$} & \plus{+2.91} & \plus{+0.76} & \minus{$-$0.54} & \minus{$-$0.93} & \plus{+3.67} & \minus{$-$5.95} & \plus{+21.81} & \plus{+19.10} & \plus{+5.09} & \minus{$-$13.92} & \plus{+8.20} & \plus{+7.78} \\
\bottomrule
\end{tabular}%
}
\caption{\small Oracle test set performance across datasets and shot settings. The best SSL baseline engineered by humans is shown in \human{yellow} per dataset/shot. The best agent-generated result is shown in \agent{purple}. We see that the best agent solution (\agent{$\text{A}_{\text{best}}$}) is generally comparable to the best SSL human baseline (\human{$\text{H}_{\text{best}}$}) and outperforms it on 8 of 12 datasets. \textbf{N} (noisy validation) and \textbf{U} (unsupervised) denote the feedback metric.}
\vspace{-1em}
\label{tab:main_results}
\end{table*}

%% file: tables/2_unsup_regret.tex
\begin{table*}[t]
\centering
\footnotesize
\setlength{\tabcolsep}{2.6pt}
\renewcommand{\arraystretch}{1.05}
% --- local colors (scoped to this table) ---
\definecolor{regretA}{HTML}{D98C8C}% 0--5
\definecolor{regretB}{HTML}{CC5555}% 5--15
\definecolor{regretC}{HTML}{B22222}% 15--30
\definecolor{regretD}{HTML}{7A0000}% 30+
\def\rgA#1{\textcolor{regretA}{#1}}%
\def\rgB#1{\textcolor{regretB}{#1}}%
\def\rgC#1{\textcolor{regretC}{#1}}%
\def\rgD#1{\textcolor{regretD}{#1}}%
\resizebox{0.97\textwidth}{!}{%
\begin{tabular}{@{}l*{12}{c}@{}}
\toprule
  & \multicolumn{2}{c}{\textbf{DTD}} & \multicolumn{2}{c}{\textbf{SUN397}} & \multicolumn{2}{c}{\textbf{RESISC45}} & \multicolumn{2}{c}{\textbf{RETINO}} & \multicolumn{2}{c}{\textbf{CLEVR}} & \multicolumn{2}{c}{\textbf{KITTI}} \\
\cmidrule(lr){2-3}\cmidrule(lr){4-5}\cmidrule(lr){6-7}\cmidrule(lr){8-9}\cmidrule(lr){10-11}\cmidrule(lr){12-13}
\textbf{Method} & 3 & 6 & 3 & 6 & 1 & 2 & 40 & 80 & 10 & 20 & 5 & 10 \\
\midrule
$\mathrm{Human}_{\mathrm{all}}$
  & \rgA{$-$3.39} & \rgA{$-$2.87} & \rgA{$-$4.69} & \rgA{$-$1.69} & \rgB{$-$14.63} & \rgA{$-$1.93} & \rgC{$-$23.92} & \rgD{$-$42.66} & \rgC{$-$15.59} & \rgC{$-$20.28} & \rgB{$-$9.10} & \rgC{$-$17.53} \\
$\mathrm{AgentSSL}^{\textbf{N}}$
  & \rgA{$-$0.67} & \rgA{$-$2.34} & \rgA{$-$0.13} & \rgA{$-$0.13} & -- & \rgB{$-$10.38} & \rgD{$-$38.12} & \rgD{$-$47.29} & \rgA{0.00} & \rgA{$-$2.66} & \rgB{$-$8.20} & \rgC{$-$19.83} \\
$\mathrm{AgentSSL}^{\textbf{U}}$
  & \rgA{$-$3.81} & \rgB{$-$10.74} & \rgA{$-$1.90} & \rgA{$-$3.60} & \rgB{$-$14.43} & \rgA{$-$2.25} & \rgC{$-$24.12} & \rgC{$-$18.62} & \rgB{$-$13.56} & \rgA{$-$3.23} & \rgB{$-$10.50} & \rgC{$-$20.96} \\
\bottomrule
\end{tabular}%
}
\caption{Unsupervised selection regret averaged across seeds. Human row selects across all human algorithms and hyperparameters using the \textbf{rank} \cite{zhang2025revisiting} metric. Agent rows select across all programs from a single trajectory using the feedback metric. Darker red denotes larger selection regret.}
\vspace{-1em}
\label{tab:unsup_regret}
\end{table*}

%% file: tables/3_active_selection_shotbridge.tex
\definecolor{lightred}{HTML}{DFA0A0}
\definecolor{darkred}{HTML}{7F1D1D}
\definecolor{lightgreen}{HTML}{8BC6A3}
\definecolor{darkgreen}{HTML}{0B5D32}

\begin{table*}[h]
\centering
\small
\setlength{\tabcolsep}{5pt}
\renewcommand{\arraystretch}{1.05}
\begin{tabular}{l|cccccc}
\toprule
\textbf{Method} & \textbf{DTD} & \textbf{SUN397} & \textbf{RESISC45} & \textbf{RETINO} & \textbf{CLEVR} & \textbf{KITTI} \\
\midrule
ActiveTesting
& \textcolor{lightred!92!darkred}{-2.35}
& \textcolor{lightred!96!darkred}{-1.17}
& \textcolor{lightred!91!darkred}{-2.61}
& \textcolor{lightgreen!2!darkgreen}{+25.59}
& \textcolor{lightred!81!darkred}{-5.41}
& \textcolor{lightgreen!60!darkgreen}{+10.46} \\

VMA
& \textcolor{lightred!94!darkred}{-1.86}
& \textcolor{lightred!95!darkred}{-1.47}
& \textcolor{lightred!74!darkred}{-7.68}
& \textcolor{lightgreen!0!darkgreen}{+26.07}
& \textcolor{lightred!85!darkred}{-4.49}
& \textcolor{lightgreen!45!darkgreen}{+14.43} \\

CODA
& \textcolor{lightred!91!darkred}{-2.50}
& \textcolor{lightred!93!darkred}{-2.09}
& \textcolor{lightred!97!darkred}{-0.87}
& \textcolor{lightgreen!19!darkgreen}{+21.22}
& \textcolor{lightred!0!darkred}{-29.09}
& \textcolor{lightgreen!44!darkgreen}{+14.63} \\

IID
& \textcolor{lightred!84!darkred}{-4.71}
& \textcolor{lightred!97!darkred}{-0.83}
& \textcolor{lightred!88!darkred}{-3.47}
& \textcolor{lightgreen!0!darkgreen}{+26.07}
& \textcolor{lightred!80!darkred}{-5.75}
& \textcolor{lightgreen!62!darkgreen}{+10.01} \\

ModelSelector
& \textcolor{lightred!92!darkred}{-2.32}
& \textcolor{lightred!97!darkred}{-0.87}
& \textcolor{lightred!55!darkred}{-13.21}
& \textcolor{lightgreen!28!darkgreen}{+18.70}
& \textcolor{lightred!93!darkred}{-2.05}
& \textcolor{lightgreen!100!darkgreen}{+0.00} \\

\bottomrule
\end{tabular}
\caption{Comparison in final selection accuracy when using labels for training versus selecting on AgentSSL unsupervised metric runs. Each cell reports the mean across active-selection seeds for train (50\%) / active selection (50\%) minus train (100\%) / unsupervised selection (0\%). IID, ActiveTesting, VMA and ModelSelector are averaged over 10 runs; CODA is deterministic.}
\label{tab:active_selection_shotbridge}
\end{table*}

%% file: sec/6_analysis.tex
\section{Analysis}

\subsection{Can Agents Leverage Unlabeled Data?}

\begin{figure}[t]
    \centering
    \includegraphics[
        width=0.78\linewidth,
        trim={0.2in 0.15in 0.15in 0in},
        clip
    ]{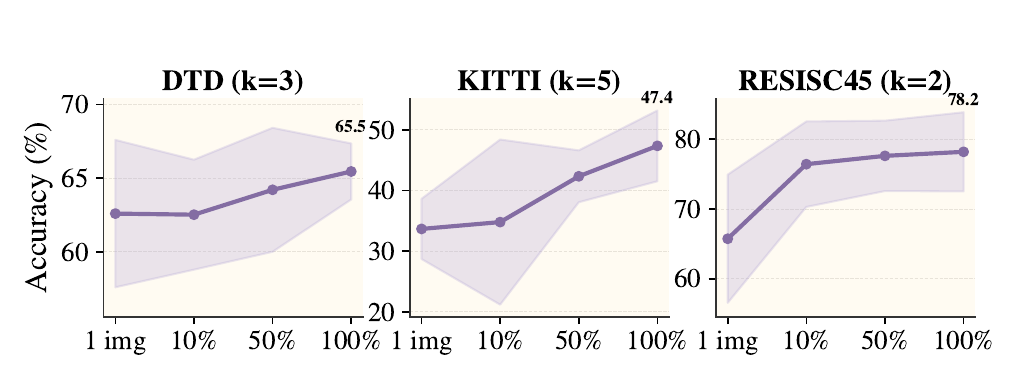}
    \caption{\small
    Performance of AgentSSL programs as the amount of unlabeled data varies while the labeled set remains fixed.
    }
    \label{fig:varying_unlabel_data}
    \vspace{-0.8em}
\end{figure}

We first study whether the discovered SSL programs actually make use of unlabeled data. To do this, we fix the top-performing program discovered by AgentSSL and vary only the size of the unlabeled corpus at evaluation time. Figure~\ref{fig:varying_unlabel_data} shows that performance generally improves as more unlabeled data becomes available while keeping the labeled set fixed. This suggests that the discovered programs rely on structure present in the unlabeled data itself.

\subsection{How Do Agents Use Unlabeled Data?}

To qualitatively analyze the agent's search space at scale, we use an LLM-as-judge pipeline to categorize the SSL strategies implemented across generated programs \cite{zhou2025llm,tong2024codejudge}. Our goal is to characterize which methods agents discover and how broadly they explore the SSL design space. We define high-level SSL categories using six classical groups from \cite{zhu2005semisupervised} and four modern deep SSL groups from \cite{Yang_2023}. Category definitions, prompts, and implementation details are provided in Appendix~\ref{appendix:llm_as_judge_details}; all LLM-as-judge-derived claims are manually verified.

\begin{figure}[t]
    \centering

    \begin{subfigure}[t]{0.48\linewidth}
        \centering
        \includegraphics[width=\linewidth]
        {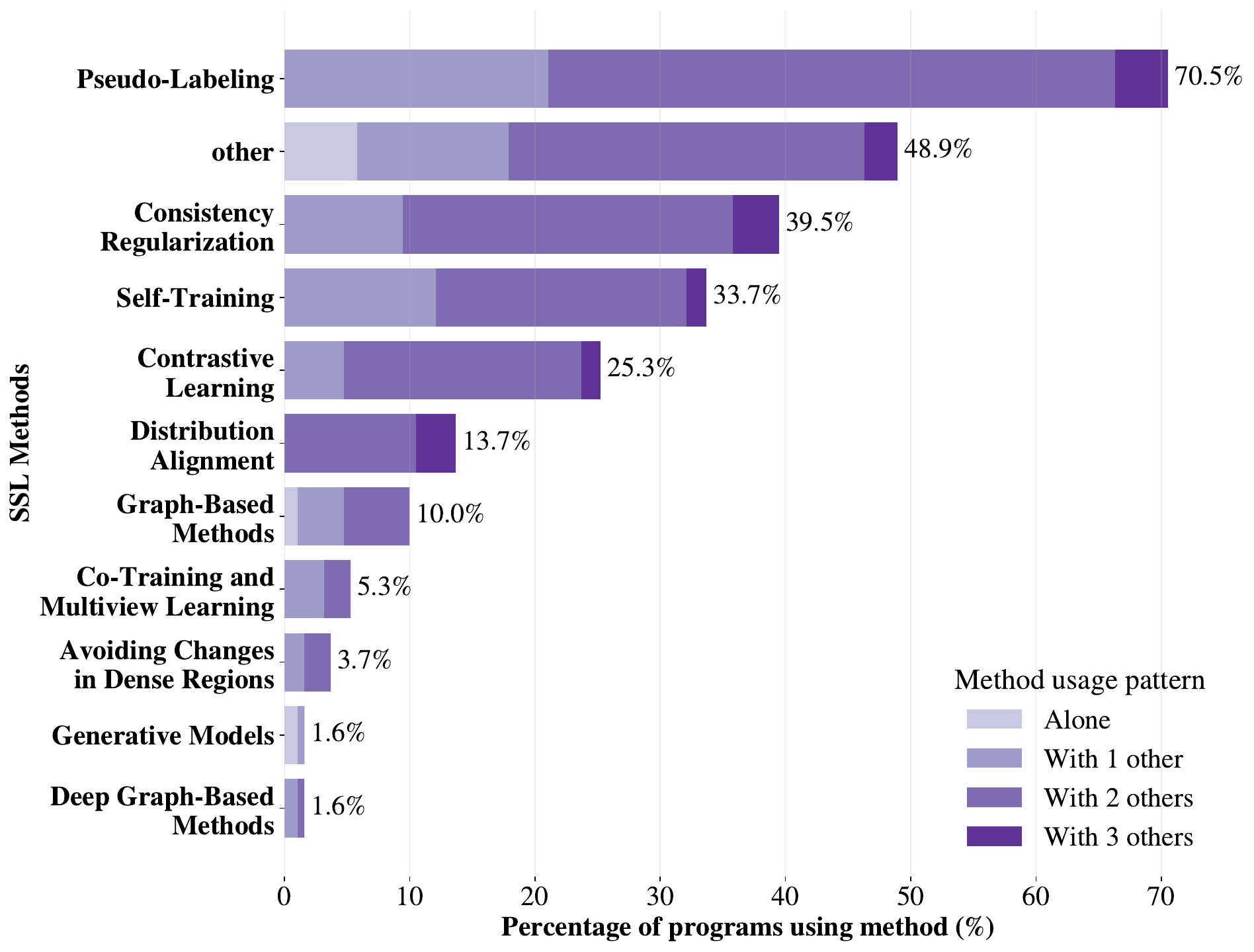}
    \label{fig:SSL_method_distribution_dtd}
    \end{subfigure}
    \hfill
    \begin{subfigure}[t]{0.48\linewidth}
        \centering
        \includegraphics[width=\linewidth]
    {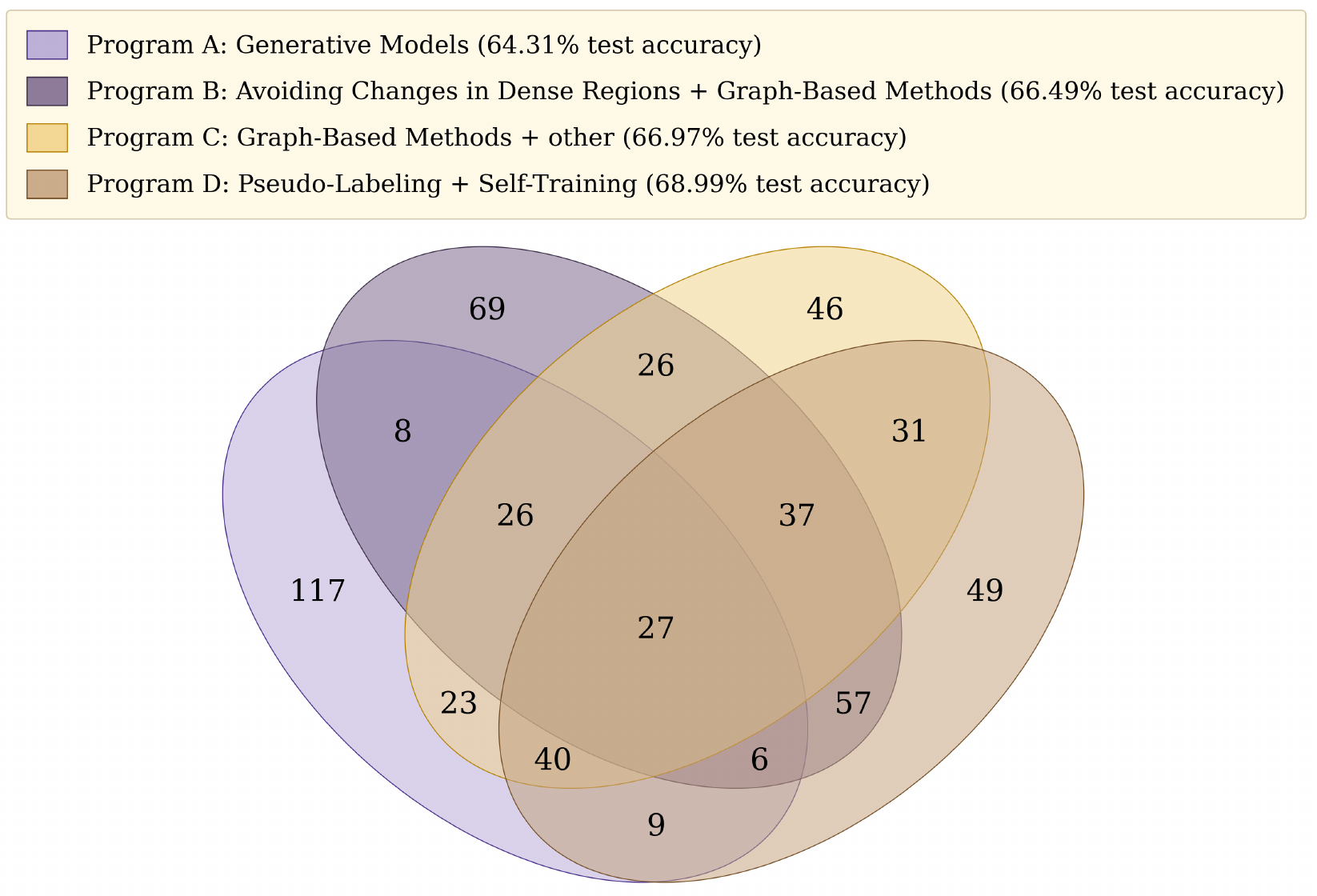}
        \label{fig:venn_diagram}
    \end{subfigure}

    \caption{\small
    \textbf{Agents discover diverse and complementary SSL strategies.}
    Left: The distribution of SSL method categories from evolutionary search rollouts on DTD.
    Right: Different high-performing programs produce confident pseudolabels on complementary subsets of the unlabeled data.
    }
    \label{fig:ssl_strategy_analysis}
\end{figure}

The left panel of Figure~\ref{fig:ssl_strategy_analysis} shows that agents explore a broad range of SSL paradigms and often combine multiple strategies within the same program. Although the frequency of individual strategies varies across datasets and search strategies, diverse and hybrid exploration is consistent across settings; full distributions are provided in Appendix~\ref{appendix:high_level_method_distributions}. We further find that these programs can be complementary. Inspired by \cite{zhang2025revisiting}, the right panel of Figure~\ref{fig:ssl_strategy_analysis} shows that high-performing programs based on different SSL strategies assign high-confidence pseudolabels to different subsets of unlabeled images. This suggests that future AgentSSL systems could use discovered programs not only as search candidates, but also as complementary pseudolabel generators.

\subsection{Do Agents Adapt SSL to Different Domains Effectively?}

\begin{figure}[t]
    \centering                               
    \includegraphics[
        width=0.78\linewidth,
    ]{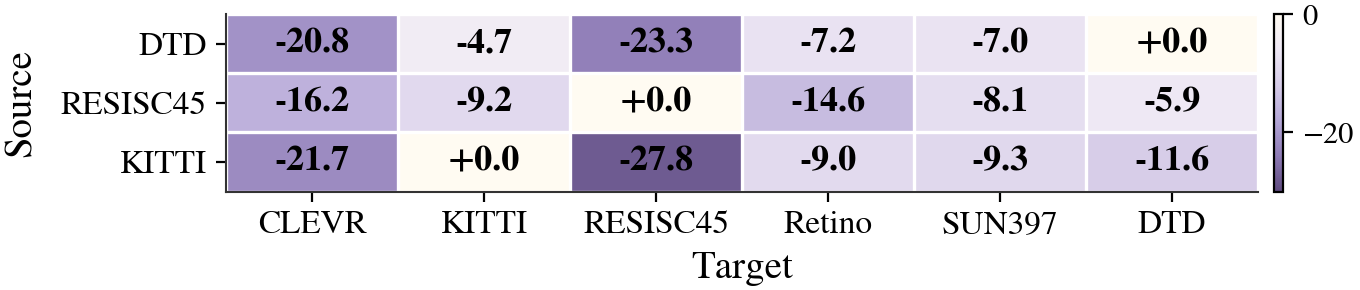}    
    \vspace{-0.5em}
    \caption{\small
    Cross-dataset generalization of agent-discovered SSL programs. Rows indicate the source dataset from which the best program is selected, and columns indicate the target dataset. Values show the change in accuracy relative to in-domain performance.
    }
    \label{fig:dataset_generalization}
    \vspace{-0.8em}
\end{figure}

One benefit of AgentSSL is the ability to discover dataset-specific learning strategies rather than relying on a single fixed SSL recipe. Figure~\ref{fig:dataset_generalization} shows that programs discovered for one dataset often transfer poorly to others i.e., the strongest programs are adapted to their source domain.

A particularly illustrative example appears on \textbf{Retinopathy} (Appendix~\ref{appendix:retino_best_program}). The best program does not train a classifier; instead, it combines frozen DINOv2 features, nearest-neighbor evidence, class prototypes, and an explicit ordinal bias. These choices align naturally with diabetic retinopathy, where labels represent ordered disease severity levels and images from similar severity stages may cluster in feature space. Despite avoiding parametric training, this program improves over the next-best baseline by 19\%, illustrating how AgentSSL can discover domain-specific inductive biases rather than simply optimizing a generic SSL recipe.

%% file: sec/7_ablations.tex
\section{Ablation}
\vspace{-0.5em}

\input{tables/4_llm_ablation}

We analyze how different agent design choices affect behavior in AgentSSL. Specifically, we study two axes: (1) the underlying language model and (2) the search strategy. We evaluate each configuration using both final test accuracy and a diversity metric based on Jaccard dissimilarity \cite{xuefei2025simpleagentsoutperformexperts} using the LLM-as-judge derived method groups. A higher diversity score indicates that the agent explores a broader range of SSL strategies and implementation choices.
Across experiments, we observe: \textit{different agent configurations often achieve comparable accuracy, but differ substantially in the diversity of SSL strategies they explore.} This suggests that agent design choices primarily influence how the search space is traversed rather than simply determining the best final program.

\textbf{LLM Backbone.}
We compare GPT-5.4 and Gemini-3.1 as the underlying program-generation models (Table~\ref{tab:llm_ablation}). Both models achieve comparable accuracy across datasets, with GPT-5.4 slightly outperforming Gemini-3.1. However, Gemini-3.1 consistently produces higher diversity scores. This mirrors the trend: stronger exploration does not always translate into higher accuracy under constrained search budgets, but substantially changes the distribution of discovered SSL solutions.

\textbf{Search Strategy.}
We compare greedy, MCTS, and evolutionary search across DTD, KITTI, and RESISC45 (Table~\ref{tab:search_strategy}). While all three strategies achieve similar accuracy ranges, they differ substantially in diversity. Evolutionary consistently produces the highest diversity scores, suggesting that population-based search encourages broader exploration. In contrast, greedy converges to narrower regions of the search space despite competitive accuracy, while MCTS lies between. These results suggest that under fixed compute budgets, the primary effect of the search strategy is not necessarily improving peak performance, but altering the breadth of SSL strategies explored during search.

\input{tables/5_search_strategy}

%% file: tables/4_llm_ablation.tex
\begin{wraptable}{r}{0.58\linewidth}
\vspace{-2.5 em}
\centering
\footnotesize
\setlength{\tabcolsep}{3pt}
\renewcommand{\arraystretch}{1.12}

\begin{tabular}{l|cc|cc|cc}
\toprule
&
\multicolumn{2}{c|}{\textbf{DTD (k=6)}} &
\multicolumn{2}{c|}{\textbf{RESISC45 (k=2)}} &
\multicolumn{2}{c}{\textbf{KITTI (k=5)}} \\
Method
& \textbf{A} & \textbf{D}
& \textbf{A} & \textbf{D}
& \textbf{A} & \textbf{D} \\
\midrule
GPT-5.4
& 0.72 & 0.37
& 0.78 & 0.37
& 0.47 & 0.42 \\

Gemini-3.1
& 0.71 & 0.46
& 0.76 & 0.59
& 0.46 & 0.52 \\
\bottomrule
\end{tabular}

\caption{\small LLM comparison with accuracy (A) and diversity (D).}
\label{tab:llm_ablation}

\vspace{-12pt}
\end{wraptable}

%% file: tables/5_search_strategy.tex
\begin{wraptable}{R}{0.52\linewidth}
\vspace{-1em}
\centering
\small
\setlength{\tabcolsep}{3pt}
\renewcommand{\arraystretch}{1.05}

\begin{tabular}{l|cc|cc|cc}
\toprule
 & \multicolumn{2}{c|}{\textbf{DTD}} 
 & \multicolumn{2}{c|}{\textbf{KITTI}} 
 & \multicolumn{2}{c}{\textbf{RESISC45}} \\
\cmidrule(lr){2-3}
\cmidrule(lr){4-5}
\cmidrule(lr){6-7}

\textbf{Method} 
& \textbf{A} & \textbf{D}
& \textbf{A} & \textbf{D}
& \textbf{A} & \textbf{D} \\

\midrule
MCTS        
& 0.70 & 0.57 
& 0.46 & 0.66 
& 0.82 & 0.58 \\

Evolutionary 
& 0.74 & 0.66 
& 0.48 & 0.65 
& 0.77 & 0.68 \\

Greedy      
& 0.72 & 0.37 
& 0.47 & 0.42 
& 0.78 & 0.37 \\

\bottomrule
\end{tabular}

\caption{\small Comparison of search strategies in terms of accuracy (A) and diversity (D).}
\label{tab:search_strategy}

\vspace{-1em}
\end{wraptable}

%% file: sec/8_ecology_application.tex
\section{Applying AgentSSL to Ecology Detection Tasks}
\vspace{-0.5em}
We evaluate AgentSSL on detection tasks derived from two ecological pipelines, WildFin \cite{wildfin} and BuckTales \cite{naik2024bucktalesmultiuavdataset}, where expert-trained detectors support downstream behavioral analysis and tracking for fish and antelope. We compare the ecologist baselines trained on the entire training split to solutions that only have access to 10\% of the data for WildFin and 20\% of the data for BuckTales. Additional dataset and experimental details in Appendix~\ref{appendix:ecology_details}.

The purpose of this case study is two-fold. First, we demonstrate a real-world setting where AgentSSL could significantly reduce both engineering and annotation cost. Second, we study whether the flexibility of MLE Agents enables the design of domain-specific SSL algorithms on tasks beyond classification that are significantly less studied. In a setting where it is unclear if unlabeled data is even useful, we additionally evaluate \textbf{AgentSL}, a subset of AgentSSL without access to unlabeled data, keeping the small labeled set. This comparison helps reveal whether a practical agent should exploit simpler supervised solutions or explore more uncertain strategies that leverage unlabeled data.

\begin{figure}[t]
    \centering
    \vspace{-0.1in}
    \includegraphics[width=0.50\linewidth]{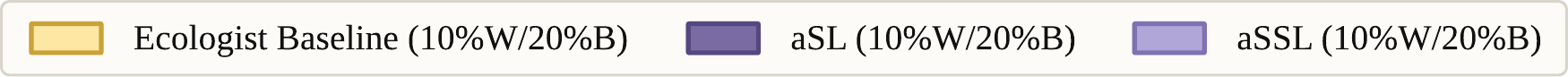}
    \includegraphics[
        width=0.65\linewidth,
        trim=0 0 0 1.0cm,
        clip
    ]{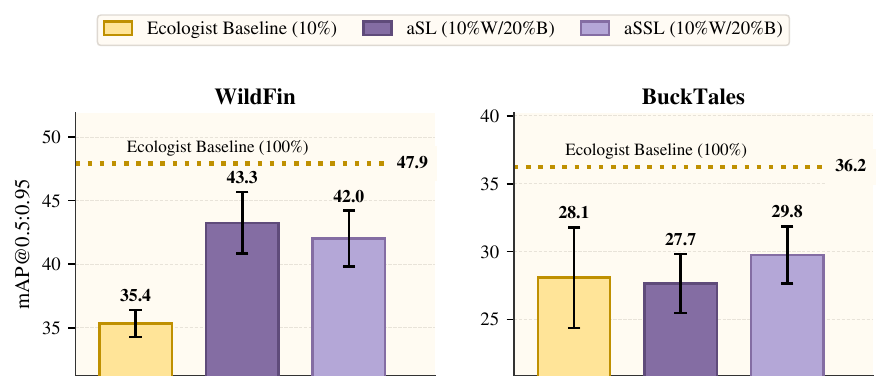}

    \vspace{-0.1in}
    \caption{\small
        Agent performance on ecological detection tasks.
        On WildFin, AgentSL and AgentSSL are comparable
        and approach the full-data expert pipeline.
        On BuckTales, both agents exhibit high variance and
        underperform the expert YOLOv8 baseline.
    }
    \vspace{-1.5em}
    \label{fig:ecological_agent_results}
\end{figure}

Figure~\ref{fig:ecological_agent_results} shows encouraging but task-dependent results. On WildFin, with just 10\% of the total data, AgentSL and AgentSSL outperform the low-label expert baseline and approach the full-data expert pipeline. On BuckTales, both agents match the low-label expert baseline. We observe that even without a large validation set, agents can discover programs that match or exceed domain-expert solutions, and on WildFin demonstrates that agentic search can substantially reduce annotation cost. 

The similar performance of AgentSL and AgentSSL across both domains also suggests that unlabeled data is not always the most valuable source of improvement. In practical settings where the utility of SSL is uncertain, strong supervised design choices may be as important as more exploratory strategies that leverage unlabeled data. Since AgentSL represents a restricted subset of the broader AgentSSL solution space, future systems should ideally search over both: exploiting simple supervised improvements when they are effective while exploring more uncertain, domain-specific uses of unlabeled data when they offer additional gains. More broadly, reducing annotation and engineering costs in scientific workflows will require agents to jointly optimize supervised design choices, SSL strategies, and domain-specific constraints.

%% file: sec/9_conclusion.tex
\vspace{-0.5em}
\section{Discussion and Conclusion}
\vspace{-0.5em}

We introduced AgentSSL, which formulates semi-supervised learning as agentic program search. Under label scarcity, MLE agents generate SSL programs that beat tuned human baselines in 8 of 12 settings when the best program is known, but neither agents nor human pipelines can reliably pick that program without labels. We also introduced \textit{ProgramSelectBench} (Appendix~\ref{appendix:selection}), which makes this selection problem measurable, and our results show that when unsupervised metrics fail, labels spent on selection buy more accuracy than the same labels spent on training.
\vspace{-0.5em}
\paragraph{Limitations and future work.} First, SSL is not always preferable to supervised learning, especially when simple modeling or engineering choices provide comparable gains \cite{oliver2019realisticevaluationdeepsemisupervised}; on both ecology tasks, AgentSL matches AgentSSL. Because our instantiation biases search toward SSL strategies, future systems should treat whether to use unlabeled data as part of the search, balancing reliable supervised improvements against more uncertain uses of unlabeled data. Second, we use AIRA-Dojo as a controlled and standardized search framework, but MLE agents are rapidly improving. Stronger LLMs and harnesses may search more efficiently and discover better solutions, so we view our results as an initial agentic baseline rather than the performance ceiling of AgentSSL. Third, our search relies on weak feedback and  measuring how search efficiency and final accuracy change as feedback gets stronger remains open. Finally, our comparison with human baselines (Table~1)  highlights the challenges of selecting programs without labels. Stronger agents will not close this gap and may widen it: the more programs an agent generates, the larger the gap can grow between the best program and the one chosen without labels. Closing it calls for unsupervised and active selection methods that stay reliable across program families, for which \textit{ProgramSelectBench} provides a testbed.
\textbf{More broadly, our results suggest that as agents make candidate programs cheap to generate, the bottleneck in AgentSSL shifts from building programs to choosing among them.}

%% file: sec/appendix.tex
\clearpage
\setcounter{page}{1}

\noindent{\large\bfseries Appendix Contents}
\vspace{0.8em}

\begingroup

\hypersetup{hidelinks}
\setlength{\parindent}{0pt}
\setlength{\parskip}{3pt}

\newcommand{\appentry}[2]{%
    \noindent
    \makebox[2em][l]{\hyperref[#1]{\ref*{#1}}}%
    \hyperref[#1]{#2}%
    \dotfill
    \pageref{#1}\par
}

\appentry{appendix:agent_details}{Agent Details}
\appentry{appendix:baseline_details}{Baseline Details}
\appentry{appendix:selection}{ProgramSelectBench: Details and Results}
\appentry{appendix:llm_as_judge_details}{LLM-as-Judge Details and Results}
\appentry{appendix:diversity}{Diversity Metric Details}
\appentry{appendix:ecology_details}{Ecology Experiment Details}
\appentry{appendix:SL_vs_SSL_app}{AgentSL vs.\ AgentSSL}
\appentry{appendix:agent_prompts}{Agent Prompts}
\appentry{appendix:agent_trajectories}{Example Agent Trajectories}
\appentry{appendix:interesting_programs}{Interesting Programs}
\appentry{appendix:societal_impact}{Broader Societal Impact}

\endgroup

\section{Agent Details}
\label{appendix:agent_details}

\subsection{Agent Hyperparameters}

The MLE Agent search hyperparameters for each strategy are provided below:
\begin{itemize}
    \item \textbf{Greedy:} 5 initial drafts, 5 improvement steps, debug probability of 1.0, and maximum debug depth of 5.
    \item \textbf{MCTS:} 5 children per expansion, UCT exploration constant of 0.25, and maximum debug depth of 20.
    \item \textbf{Evolutionary:} 5 individuals per generation, 100 generations, crossover probability of 0.5, and maximum debug depth of 10.
    \item \textbf{LLM sampling:} Draft, improve, and debug operators use temperature 0.6 and top-$p=0.95$; analysis uses temperature 0.5.
\end{itemize}

\subsection{Feedback Metric Details}
\label{appendix:feedback_metrics}

\paragraph{Noisy Validation}
\label{noisy_validation_details}

One SSL feedback metric we test is a noisy validation metric. This is created by splitting the existing limited training set into train and validation. In order to manage variance due to limited size of the validation set, we perform $k$-fold noisy validation to get better signal. However, because there is a tradeoff between reducing variance in performance estimates and managing compute time, we manually tune the $k$ hyperparameter per dataset. The exact fold number picked for each dataset in this work is listed below: 
\begin{itemize}
    \item \textbf{DTD} : 3-shot and 6-shot both have 3-fold noisy validation
    \item \textbf{RESISC45} : 1-shot is not applicable for noisy validation and 2-shot has 2-fold validation
    \item \textbf{SUN397} : 3-shot and 6-shot both have 3-fold noisy validation
    \item \textbf{Retinopathy} : 40-shot and 80-shot both have 1-fold noisy validation 
    \item \textbf{CLEVR} : 10-shot has 5-fold and 20-shot has 4-fold noisy validation
    \item \textbf{KITTI} : 5-shot and 10-shot both have 3-fold noisy validation
\end{itemize}

Automatically determining the best $k$ value for a task or even dynamically determining the resources used to reduce uncertainty in performance estimates during search is an interesting future research direction.

\paragraph{Unsupervised Metrics}
\label{unsupervised_metrics_Datils}

Inspired by prior work \cite{zhang2025revisiting}, we use an ensemble of different unsupervised metrics as feedback for the agentic search. Since different unsupervised metrics capture different properties of the learned representations and predictions, an ensemble should be able to balance these biases.

Because the metrics operate on different numerical scales, we map each score to $[0,1]$ using predefined global bounds derived from its valid range, clipping values to those bounds before normalization. We deliberately restrict the ensemble to metrics with well-defined normalization ranges. The final unsupervised fitness score is the mean of the selected normalized metrics, yielding a single scalar objective for ranking candidate programs during search.

The ensembled unsupervised fitness score includes the following metrics:
\begin{itemize}
    \item \textbf{AMI}~\cite{vinh2009information}: mutual information between cluster assignments, corrected for chance.
    \item \textbf{ARI}~\cite{Halkidi2002ClusterVM}: pairwise agreement between cluster assignments, corrected for chance.
    \item \textbf{V-measure}~\cite{rosenberg-hirschberg-2007-v}: entropy-based measure balancing cluster homogeneity and completeness.
    \item \textbf{FMI}~\cite{fowlkes1983method}: agreement between clusterings based on pairwise precision and recall.
    \item \textbf{Silhouette Score}~\cite{rousseeuw1987silhouettes}: cluster compactness relative to separation from neighboring clusters.
    \item \textbf{RankMe}~\cite{garrido2023rankmeassessingdownstreamperformance}: effective rank of learned representations as a proxy for representation quality.
    \item \textbf{BNM}~\cite{cui2020discriminabilitydiversitybatchnuclearnorm}: nuclear norm of batch predictions, capturing discriminability and prediction diversity.
    \item \textbf{SND}~\cite{saito2021tune}: local feature-space density measured through pairwise feature similarities.
\end{itemize}

\begin{figure}[h]
    \centering
    \makebox[\textwidth][c]{%
        \begin{subfigure}{0.49\textwidth}
            \centering
            \includegraphics[width=\linewidth]{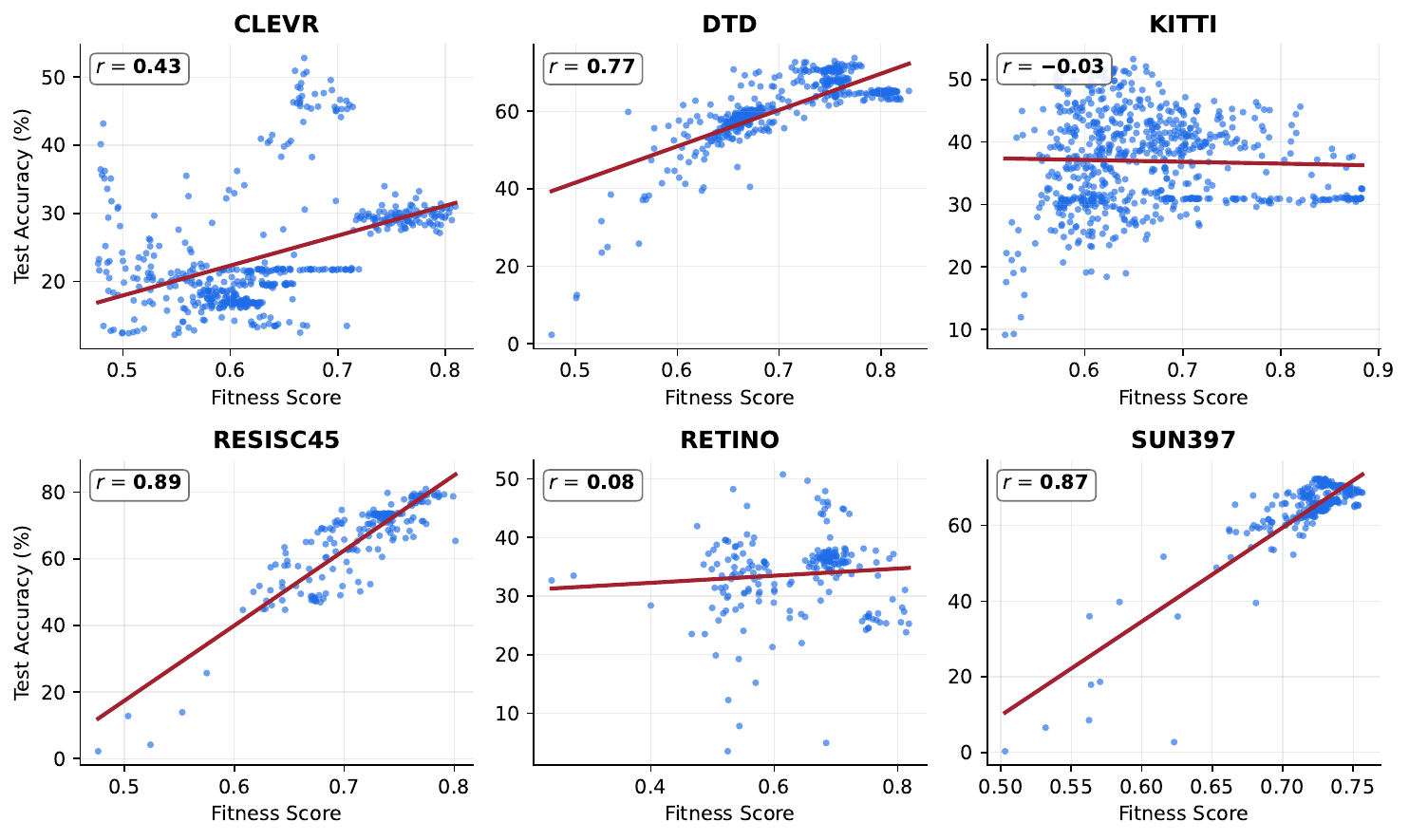}
            \caption{Unsupervised Metrics setting}
        \end{subfigure}
        \hspace{0.02\textwidth}
        \begin{subfigure}{0.49\textwidth}
            \centering
            \includegraphics[width=\linewidth]{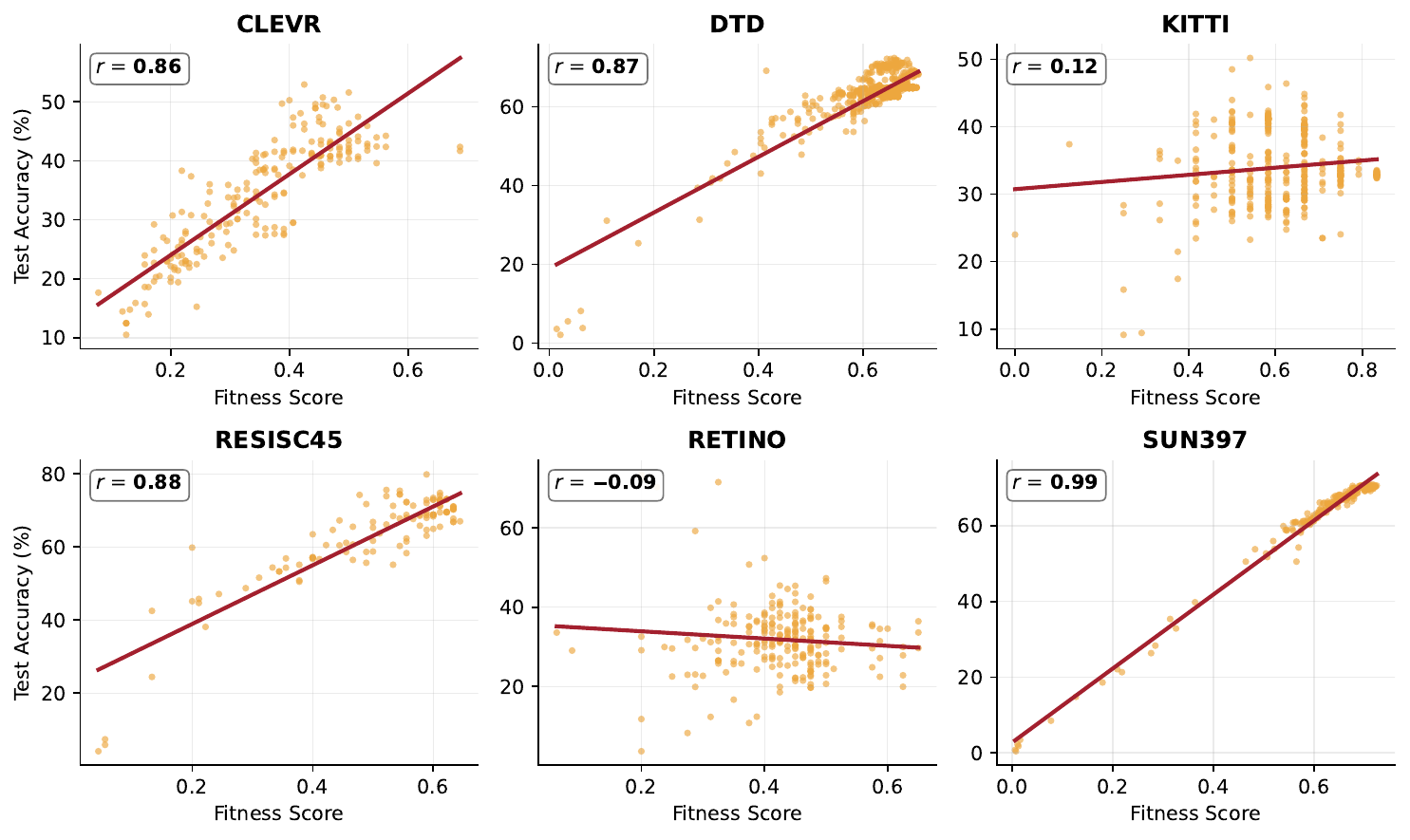}
            \caption{Noisy Validation setting}
        \end{subfigure}%
    }
    \caption{Correlation between the agent's feedback metric and test accuracy for (a) the unsupervised feedback metric and (b) noisy validation. Each point is one non-buggy program generated during search, pooled across both shot settings and all three search seeds for each dataset. The x-axis is the program's fitness score or feedback metric, the numerical score the agent received during search as feedback. The y-axis is its test accuracy averaged over three evaluation seeds. Red lines are least-squares fits, and $r$ is the Pearson correlation coefficient.}

    \label{fig:unsup_corr}
\end{figure}

These metrics have also been studied for unsupervised model selection in prior work ~\cite{ericsson2023betterpracticesdomainadaptation,zhang2025revisiting}. However, they are not always correlated to test accuracy. In Figure~\ref{fig:unsup_corr}, we see that both feedback metrics, unsupervised and noisy validation, can even be negatively correlated with test accuracy on datasets like KITTI and Retinopathy. We generally see a similar trend across all the unsupervised metrics as well as for other ensemble strategies like \textbf{rank}, which selects the program with the best average ranking across each metric \cite{zhang2025revisiting}. 

\subsection{Total Compute}
\label{appendix:agent_compute}
We scale the agent search budget according to the computational cost of the corresponding SSL baselines, since candidate evaluation time increases substantially with dataset size. For the smaller datasets (DTD, RESISC45, and KITTI), each agent run is allocated 8 GPU hours on an NVIDIA RTX 6000 Ada Generation GPU. For the larger datasets, each run is allocated 15 GPU hours on either an NVIDIA H100 GPU or an NVIDIA RTX 6000 Blackwell Max-Q GPU.

Across 6 datasets, 2 shot settings, and 3 random seeds, each agent setting comprises 36 runs: 18 with an 8-hour budget and 18 with a 15-hour budget. We evaluate two agent settings, unsupervised and noisy validation, for a total compute budget of
$
2 \times (18 \times 8 + 18 \times 15) = 828
$
GPU hours. This number represents the amount of compute time taken to perform agent trajectories on a single seed. In our evaluation, we also compute every program on 2 additional seeds which triples the total compute cost to 2,484 GPU hours. 

\section{Baseline Details}
\label{appendix:baseline_details}
\subsection{Methods and Hyperparameters}

We evaluate the same supervised and SSL baselines described in the main text. For all baselines, we follow the training protocol and hyperparameter search space from prior work~\cite{zhang2025revisiting}, using the authors' released implementation. We consider configurations based on CLIP ViT-B/16~\cite{radford2021learningtransferablevisualmodels} and DINOv2-B~\cite{oquab2024dinov2learningrobustvisual}, with parameter-efficient fine-tuning using LoRA~\cite{hu2021loralowrankadaptationlarge}, and reproduce the prescribed hyperparameter tuning procedure.

To reduce sensitivity to the particular labeled examples selected in the few-shot regime, we rerun every baseline across three independently sampled labeled/unlabeled splits for each dataset and shot setting. The original study reports results on a single labeled split. For each seed, agents and baselines are evaluated on the same labeled and unlabeled split, so comparisons are paired under identical data conditions. We report results aggregated across the three seeds.

This evaluation protocol requires a large number of trained models for each dataset and shot setting:

\begin{table}[h]
\centering
\small
\begin{tabular}{lc}
\toprule
\textbf{Baseline Method} & \textbf{Models trained} \\
\midrule
Supervised & 144 \\
FixMatch & 18 \\
SoftMatch & 18 \\
FineSSL & 18 \\
PET & 6  \\
V-PET & 6  \\
\bottomrule
\end{tabular}
\vspace{1em}
\caption{Number of model configurations evaluated per dataset and shot setting. PET and V-PET additionally require predictions from 144 LoRA supervised and 144 AdaptFormer~\cite{chen2022adaptformeradaptingvisiontransformers} supervised models.}
\label{tab:baseline_model_counts}
\end{table}

Accounting for all datasets, shot settings, random splits, and required pretrained/adapted models, we train \textbf{2,520 models for baseline results} and \textbf{4,248 models in total} when including the additional AdaptFormer supervised runs required to compute PET and V-PET.

\subsection{Total Compute}
\label{appendix:baseline_compute}
The breadth of baseline algorithms and hyperparameter configurations makes the baseline evaluation substantially more computationally expensive than any individual AgentSSL search. As seen in Table~\ref{tab:baseline_runtime_totals}, the total compute used to train and tune the baseline models far exceeds the total compute allocated to the agent experiments. Note that PET and V-PET include several other computationally expensive steps that we do not include.

\input{tables/7_baseline_runtime}
\begin{table*}[t]
\centering
\resizebox{\textwidth}{!}{%
\begin{tabular}{*{14}{c}}
\toprule
& \multicolumn{2}{c}{CLEVR}
& \multicolumn{2}{c}{DTD}
& \multicolumn{2}{c}{KITTI}
& \multicolumn{2}{c}{RESISC45}
& \multicolumn{2}{c}{RETINO}
& \multicolumn{2}{c}{SUN397}
& \multirow{2}{*}{\textbf{TOTAL}} \\
\cmidrule(lr){2-3}
\cmidrule(lr){4-5}
\cmidrule(lr){6-7}
\cmidrule(lr){8-9}
\cmidrule(lr){10-11}
\cmidrule(lr){12-13}
& 10 & 20
& 3 & 6
& 5 & 10
& 1 & 2
& 40 & 80
& 3 & 6
& \\
\midrule
Baseline
& 354 & 354
& 354 & 354
& 354 & 354
& 354 & 354
& 354 & 354
& 354 & 354
& 4,248 \\
Noisy-Val
& 375 & 243
& 1,401 & 498
& 603 & 537
& -- & 348
& 195 & 477
& 375 & 309
& 5,361 \\
Unsupervised
& 1,113 & 456
& 1,251 & 309
& 798 & 1,317
& 333 & 261
& 372 & 237
& 537 & 384
& 7,368 \\
\midrule
\textbf{Total}
& 1,842 & 1,053
& 3,006 & 1,161
& 1,755 & 2,208
& 687 & 963
& 921 & 1,068
& 1,266 & 1,047
& \textbf{16,977} \\
\bottomrule
\end{tabular}%
}
\caption{ProgramSelectBench tasks and number of programs in each task.}
\label{appendix:evaluation-counts}
\end{table*}

\section{ProgramSelectBench: Details and Results}
\label{appendix:selection}
We compile and release \textbf{ProgramSelectBench}, a precomputed benchmark for unsupervised and active program selection consisting of 16,977 evaluated program instances across six classification datasets. ProgramSelectBench is constructed from the programs evaluated in AgentSSL, including both agent-generated programs and human-designed baselines. The benchmark provides a controlled setting for studying program selection independently of program discovery, allowing future work to evaluate selection strategies without the computational cost of generating and executing new programs. The benchmark is publicly available at \url{https://huggingface.co/datasets/ethan-y-lin/ProgramSelectBench}.

\subsection{Selection task definition} A selection task consists of a pool of candidate programs \(\mathcal{H}=\{h_1,\ldots,h_K\}\), evaluated on a common set of examples. The objective is to identify the program with the highest ground-truth accuracy without directly observing the full set of ground-truth labels. In unsupervised selection, methods may use the programs' predictions and extracted features to estimate their relative performance. In active selection, methods additionally have access to a limited annotation budget and may actively query labels to distinguish between candidate programs. Selection performance is measured by the ground-truth accuracy of the selected program, or equivalently its regret relative to the best program in the pool.

\subsection{Benchmark Details.} We organize ProgramSelectBench into 35 selection tasks corresponding to the dataset, shot, and evaluation settings considered in AgentSSL. Each task contains all evaluated program instances for the corresponding setting, including both agent-generated programs and human-designed baselines. Table~\ref{appendix:evaluation-counts} summarizes the number of programs in each task.

We treat programs trained on different labeled-data splits as distinct candidate instances. Although these instances may share the same algorithm and implementation, differences in training data can produce substantial variation in their predictions and test performance. Because each baseline configuration and agent-generated program is evaluated using three training splits, we include each trained instance as a separate candidate in the corresponding selection pool.

\begin{figure}[h]
    \centering
    \includegraphics[width=0.95\linewidth]{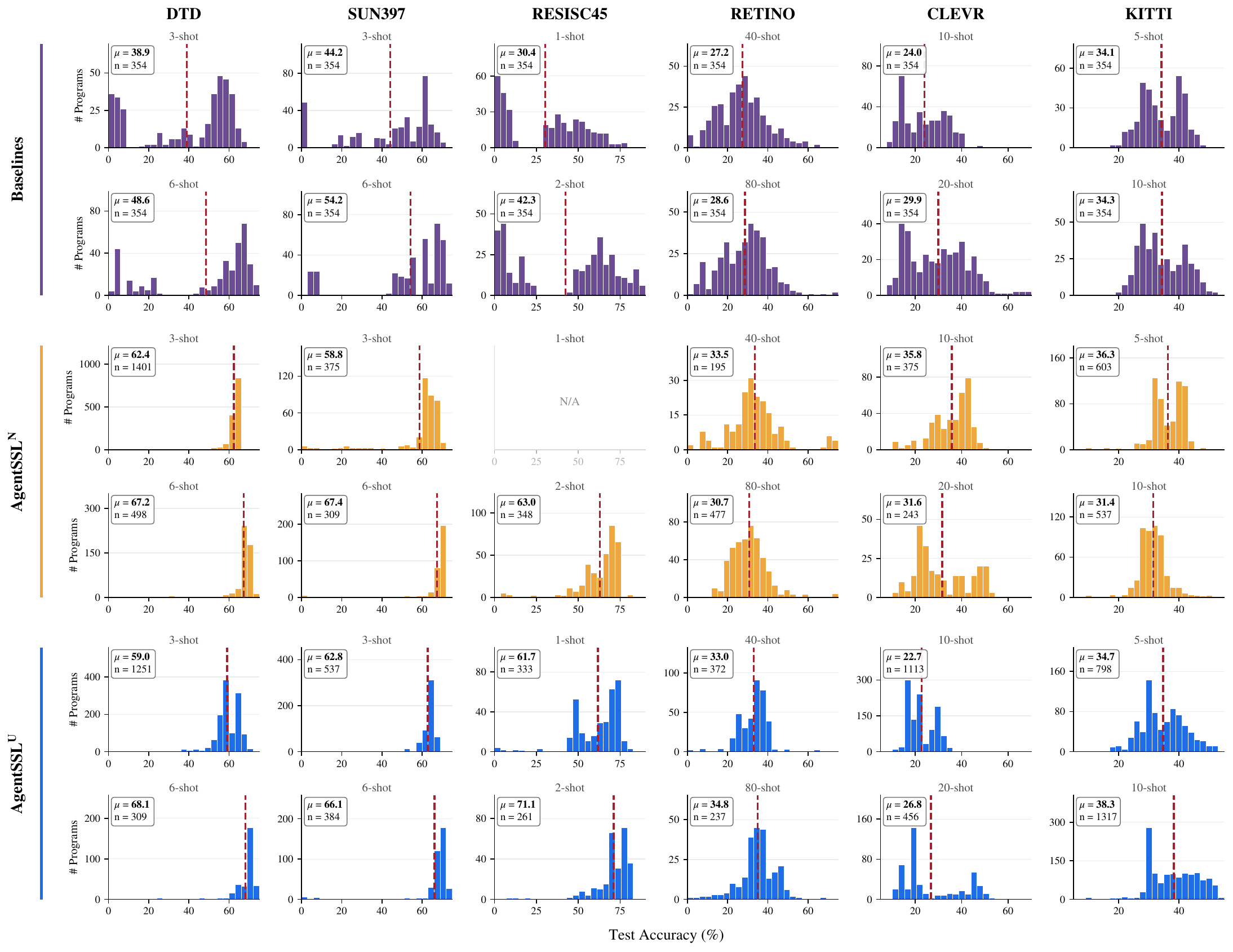}
    \caption{Accuracy Distribution for each ProgramSelectBench task.}
    \label{appendix:psb_acc_hist}
\end{figure}

Figure~\ref{appendix:psb_acc_hist} illustrates the distributions of program accuracy across the benchmark tasks. The tasks vary in both the number of candidate programs and the distribution of their performance, providing a range of selection problems. Furthermore, the benchmark is not restricted to the task partitions used in AgentSSL: any subset of programs evaluated on the same examples can define a new selection task, enabling the construction of alternative candidate pools. For example, we could construct a larger selection task on DTD by combining every shot and setting pools into a single large pool.

\textbf{Released artifacts.} ProgramSelectBench provides precomputed predictions and unsupervised evaluation metrics for all candidate programs. For each selection task, we aggregate the predictions of all programs on the corresponding test set into a single tensor file, allowing selection methods to efficiently load and compare the full candidate pool without executing individual programs. We additionally provide precomputed unsupervised metrics for each program to facilitate the evaluation of unsupervised selection methods. ProgramSelectBench also includes all agent-generated programs alongside textual descriptions of the corresponding method and implementation. We provide code for evaluating these programs and extracting their predictions and intermediate features. Similarly, we provide the code to evaluate all baseline methods as well and extract predictions and features. Although we have computed the features which are useful for unsupervised metrics, we do not include them in the benchmark release due to their substantially larger storage requirements.

\subsection{Selection Results}

For \textbf{unsupervised selection}, we evaluate all the unsupervised metrics discussed in Section~\ref{appendix:feedback_metrics} on ProgramSelectBench. 
For \textbf{active selection},
we evaluate several active selection methods including ActiveTesting \cite{kossen2021activetestingsampleefficientmodel}, VMA \cite{matsuura2023active}, CODA \cite{kay2025consensusdrivenactivemodelselection}, ModelSelector \cite{okanovic2024modelswrongusefulmodel}, and an IID baseline. We evaluate each one for ten random seeds (except for CODA which is deterministic), taking the average performance across seeds, and
use default hyperparameters. The number of iterations, or label budget, provided for each active selection algorithm and task is defined by: 
$$\text{min}(25, \text{\# of classes} \times \text{\# of training images per class})$$

\begin{figure*}[h]
    \centering

    \makebox[\textwidth][c]{%
        \includegraphics[width=0.5\textwidth]{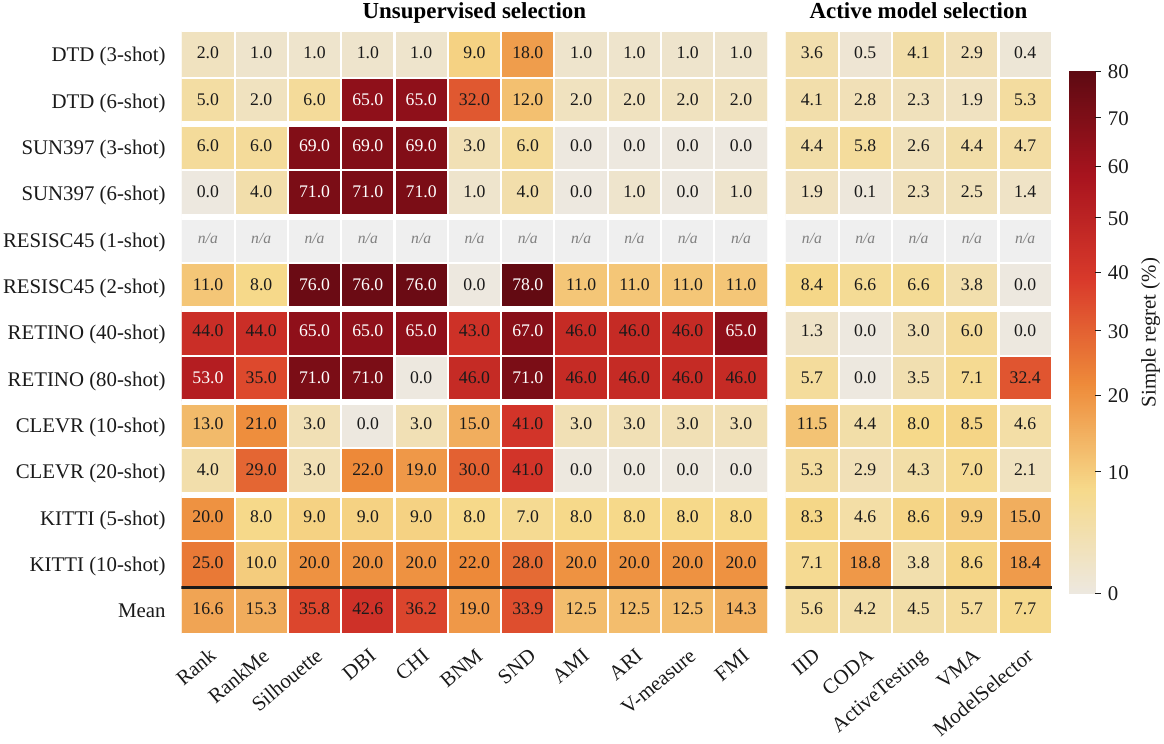}
        \hspace{0.01\textwidth}
        \includegraphics[width=0.5\textwidth]{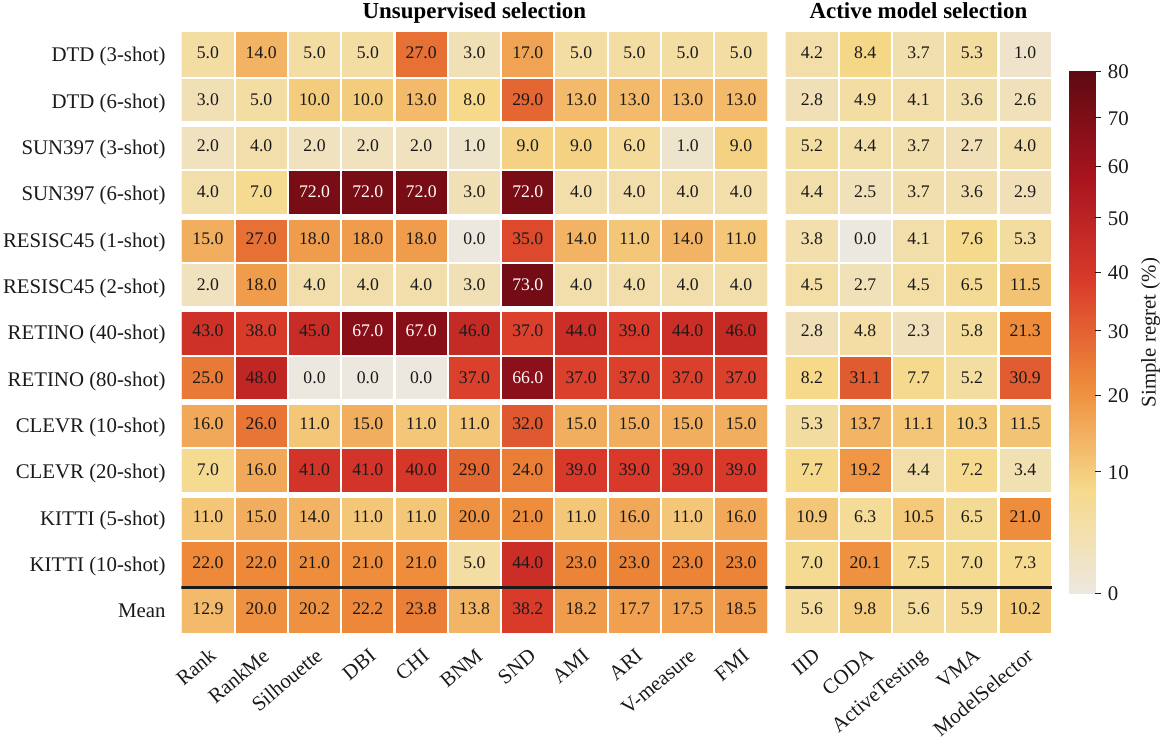}
    }
    \makebox[\textwidth][c]{%
        \includegraphics[width=0.5\textwidth]{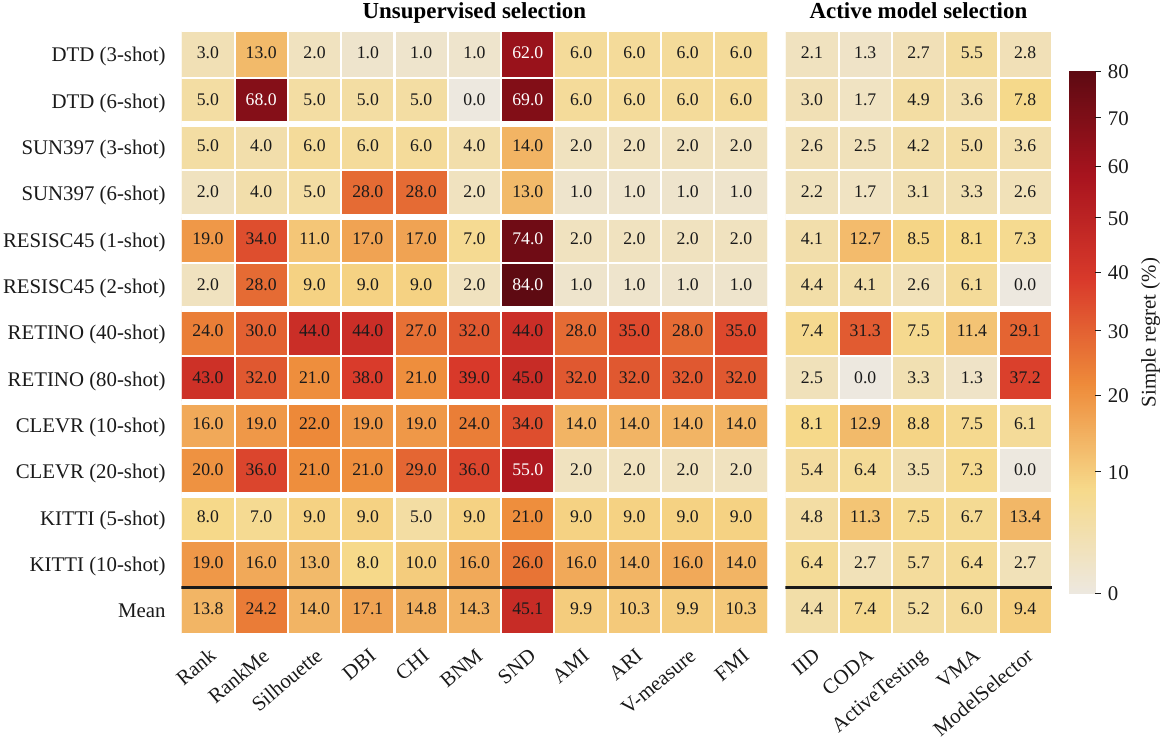}
    } 

    \caption{\small Selection regret for both unsupervised selection metrics and active selection metrics across datasets and shot counts, for agent noisy validation tasks (left), agent unsupervised tasks (right), and baseline tasks (bottom).}
    \label{appendix:simple-regret}
\end{figure*}

The main selection results are shown in Figure~\ref{appendix:simple-regret}. Similar to the results in Table~\ref{tab:unsup_regret}, individual unsupervised metrics generally perform very poorly on all Retinopathy and KITTI tasks with some exceptions. We also see that even when dataset and shot are fixed, there is still high variance in the performance of unsupervised metrics across different program pools. For example, AMI, ARI, and FMI perform well on CLEVR (0 and 3\% regret) for agent noisy validation tasks, however, on agent unsupervised tasks, they perform very poorly (15\% and 39\%). This inconsistency across datasets and even program pools suggests that using these metrics in practice may be unreliable. 

Active Model Selection methods, with just 25 labels or less, have much smaller simple regret than the unsupervised selection methods. Notably, the IID baseline performs very well and even has the best average simple regret on the baseline tasks. Suprisingly, we also observe that the IID baseline actually also has smaller variance than ActiveTesting and VMA (Table~\ref{appendix:active-selection-std}). 

\begin{table}[t]
\centering
\small
\begin{tabular}{lrrrrr}
\toprule
\textbf{Metric} & \textbf{IID} & \textbf{ActiveTesting} & \textbf{VMA} & \textbf{CODA} & \textbf{ModelSelector} \\
\midrule
Simple regret & 3.80 & 4.14 & 4.78 & 0.00 & 1.73 \\
Cumulative regret & 70.07 & 83.76 & 90.60 & 0.00 & 12.50 \\
\bottomrule
\end{tabular}
\caption{Mean per-task standard deviation for cumulative regret and simple regret across active-selection seeds. Simple regret is regret at the final iteration. Cumulative regret sums the simple regrets across all 25 iterations.}
\label{appendix:active-selection-std}
\end{table}

\begin{figure}
    \centering
    \includegraphics[width=1\linewidth]{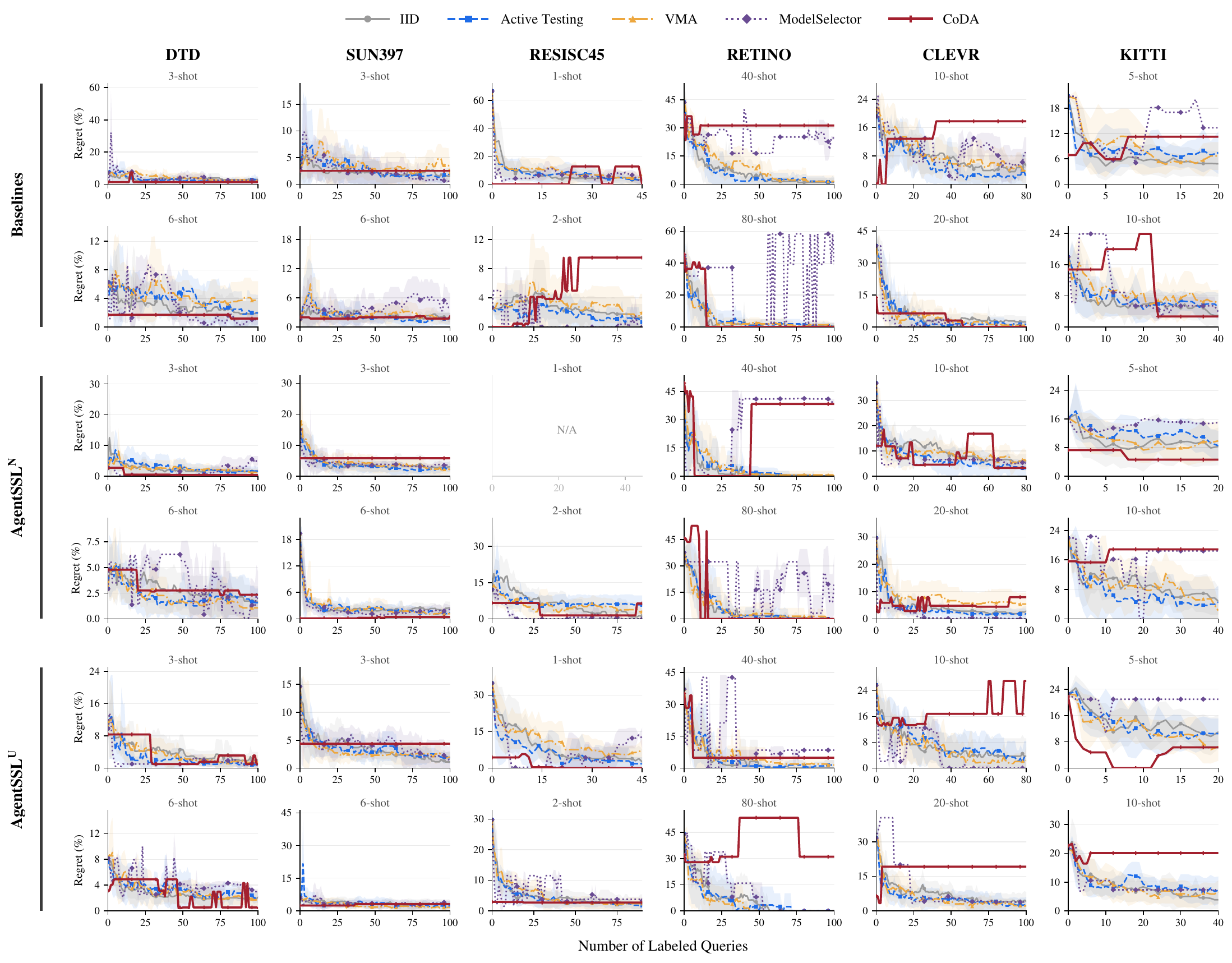}
    \caption{Regret vs. Iterations plot for all active selection methods on every ProgramSelectBench task.}
    \label{appendix:regret-vs-iteration}
\end{figure}

We analyze the active model selection results more closely in Figure~\ref{appendix:regret-vs-iteration}, where we plot the regret versus iterations where the iterations per task is defined by 
$$\text{min}(100, \text{\# of classes} \times \text{\# of training images per class}).$$ We can see that there is high variance across seeds on IID, ActiveTesting, and VMA, which lowers their real-world utility. CODA (which is deterministic) and ModelSelector exhibit much smaller variance, and often converge more quickly to zero regret, however are more unstable. For example, on the RETINO (40-shot) noisy validation task, both CODA and ModelSelector get to zero regret with less than 25 labels, but after 40 labels, they suddenly select a program with 30\% worse accuracy. 

Our benchmark results suggest that there is significant room for improving both unsupervised selection methods and active selection methods, which will greatly increase the practical performance of any data-efficient method.  

\section{LLM-as-Judge Details and Results}
\label{appendix:llm_as_judge_details}

We use an LLM-as-judge pipeline to analyze the programs produced during agent search at scale with GPT-5.4. Given a program as input, the judge extracts: (i) the models used, (ii) the high-level SSL method groups represented by the program, (iii) implementation-level details, (iv) whether the program performs training, and (v) a short rationale summarizing the previous choices. This structured representation lets us quantify which parts of the SSL design space agents explore, while preserving enough detail for manual validation. We show the high level method distribution results in this paper and will release all analysis output data and agent-generated programs. 

\subsection{High-Level SSL Method Groups}
\label{appendix:llm_as_judge_method_groups}

We categorize each program using a set of classical and modern SSL method groups. The classical groups are adapted from prior SSL surveys, which organize methods by assumptions about the data distribution, decision boundary, and relationship between labeled and unlabeled examples~\cite{zhu2005semisupervised}. The modern deep SSL groups follow recent taxonomies that distinguish deep generative methods, consistency regularization, graph-based methods, pseudo-labeling, and hybrid approaches~\cite{Yang_2023}. We additionally include contrastive learning because many modern programs use representation learning objectives over labeled and unlabeled data before or during downstream training~\cite{technologies9010002} and we see this reflected in the agent outputs.

\begin{itemize}
    \item \textbf{Generative Models.} Generative SSL methods model the joint distribution of inputs and labels, or the class-conditional data distribution, and use unlabeled data to improve estimates of the underlying data-generating process.
    
    \item \textbf{Graph-Based Methods.} Graph-based SSL methods construct a graph over labeled and unlabeled examples and propagate label information through the graph under the assumption that nearby or strongly connected examples should share labels.
    
    \item \textbf{Avoiding Changes in Dense Regions.} Low-density separation methods encourage decision boundaries to lie in low-density regions of the input or feature space, rather than cutting through clusters of unlabeled examples.
    
    \item \textbf{Distribution Alignment.} Distribution alignment methods adjust predictions, pseudo-labels, or learned representations so that the model's predicted label distribution better matches a target prior or expected class distribution.
    
    \item \textbf{Self-Training.} Self-training methods iteratively train a model on labeled data, assign labels to unlabeled examples using the model's own predictions, and retrain or refine the model using these pseudo-labeled examples.
    
    \item \textbf{Co-Training and Multiview Learning.} Co-training and multiview methods train multiple models or views of the data that exchange predictions, relying on complementary information across views to label unlabeled examples.
    
    \item \textbf{Deep Generative Models.} Deep generative SSL methods use neural generative models, such as VAEs or GANs, to learn latent structure from unlabeled data and improve supervised prediction.
    
    \item \textbf{Consistency Regularization.} Consistency regularization methods encourage a model to produce stable predictions under input perturbations, data augmentation, dropout, or other stochastic transformations.
    
    \item \textbf{Deep Graph-Based Methods.} Deep graph-based SSL methods combine neural representations with graph propagation, graph neural networks, or feature-space neighborhood structure to transfer label information across labeled and unlabeled examples.
    
    \item \textbf{Pseudo-Labeling.} Pseudo-labeling methods assign hard or soft labels to unlabeled examples using model predictions and then train on these pseudo-labeled examples, often with confidence thresholds or filtering.
    
    \item \textbf{Contrastive Learning.} Contrastive learning methods learn representations by pulling together positive pairs and pushing apart negative pairs, often using unlabeled data to improve feature quality before or during SSL.
\end{itemize}

\subsection{Prompt}
\label{appendix:llm_as_judge_prompt}

The prompt used for the LLM-as-judge pipeline is shown below.

\input{prompts/llm_as_judge_prompt}

\subsection{High-Level Method Distributions per Dataset}
\label{appendix:high_level_method_distributions}

\begin{figure}[ht]
    \centering

    \begin{subfigure}[t]{0.32\textwidth}
        \centering
        \includegraphics[width=\linewidth]{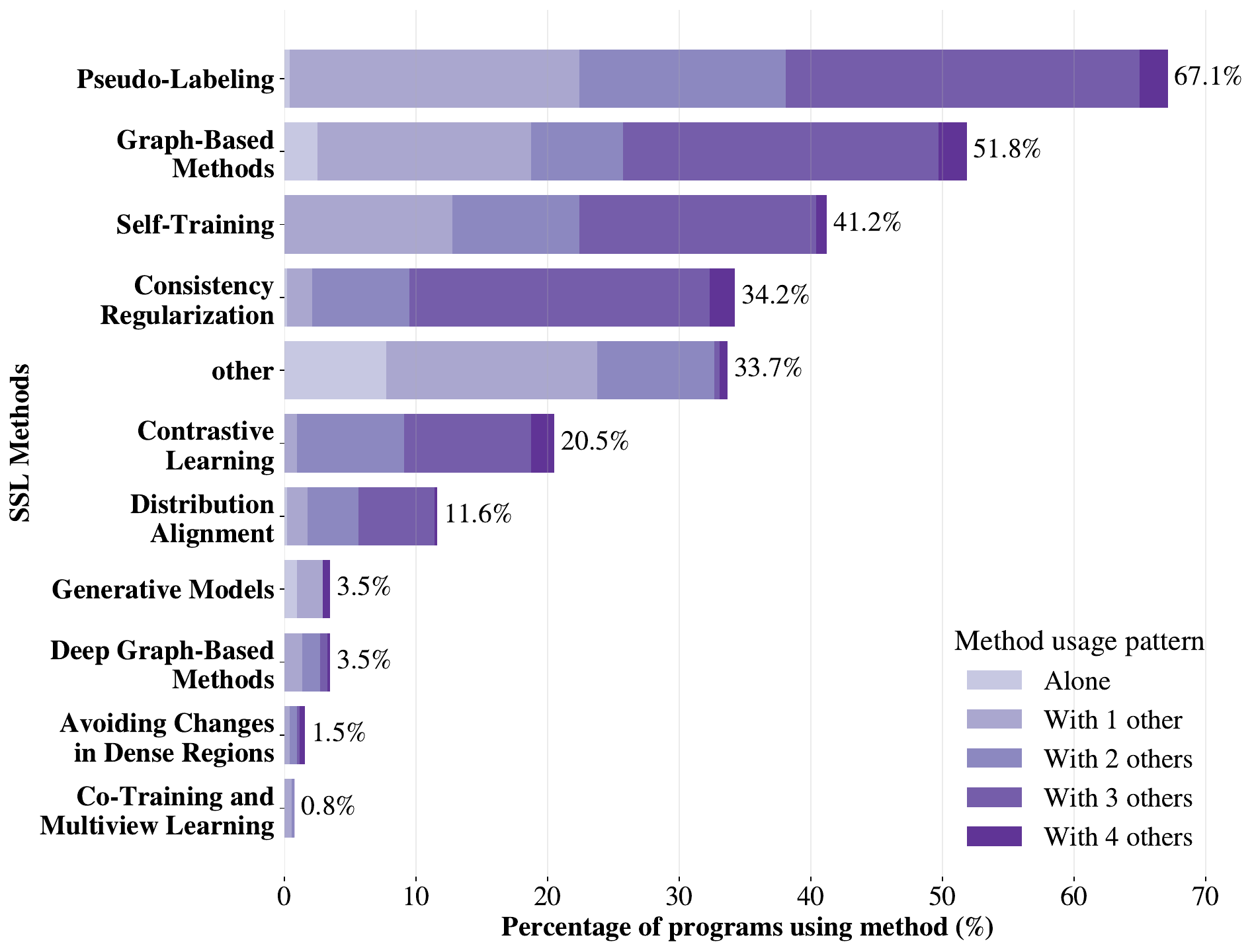}
        \caption{DTD}
        \label{fig:dtd_hlmd}
    \end{subfigure}
    \hfill
    \begin{subfigure}[t]{0.32\textwidth}
        \centering
        \includegraphics[width=\linewidth]{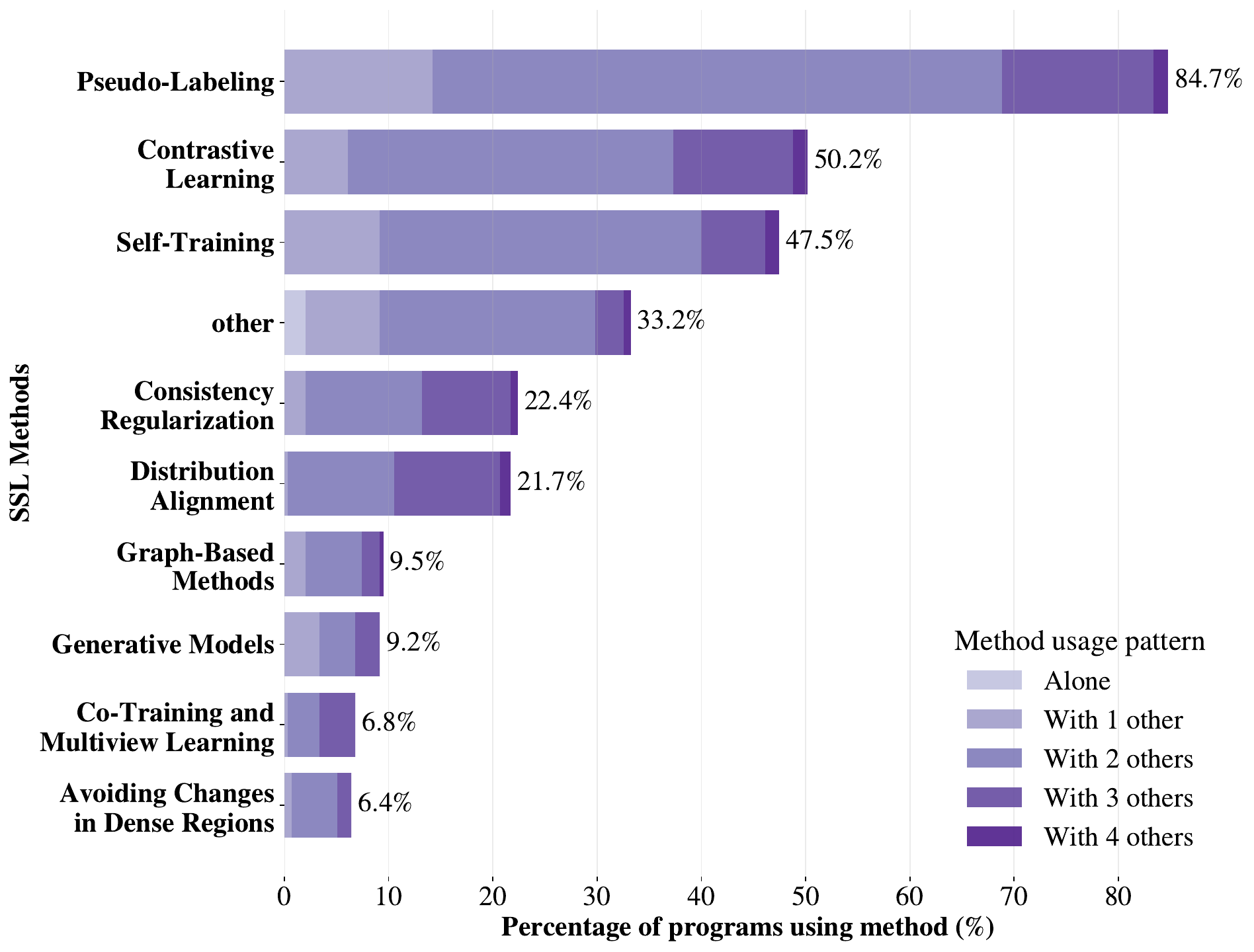}
        \caption{SUN397}
        \label{fig:sun397_hlmd}
    \end{subfigure}
    \hfill
    \begin{subfigure}[t]{0.32\textwidth}
        \centering
        \includegraphics[width=\linewidth]{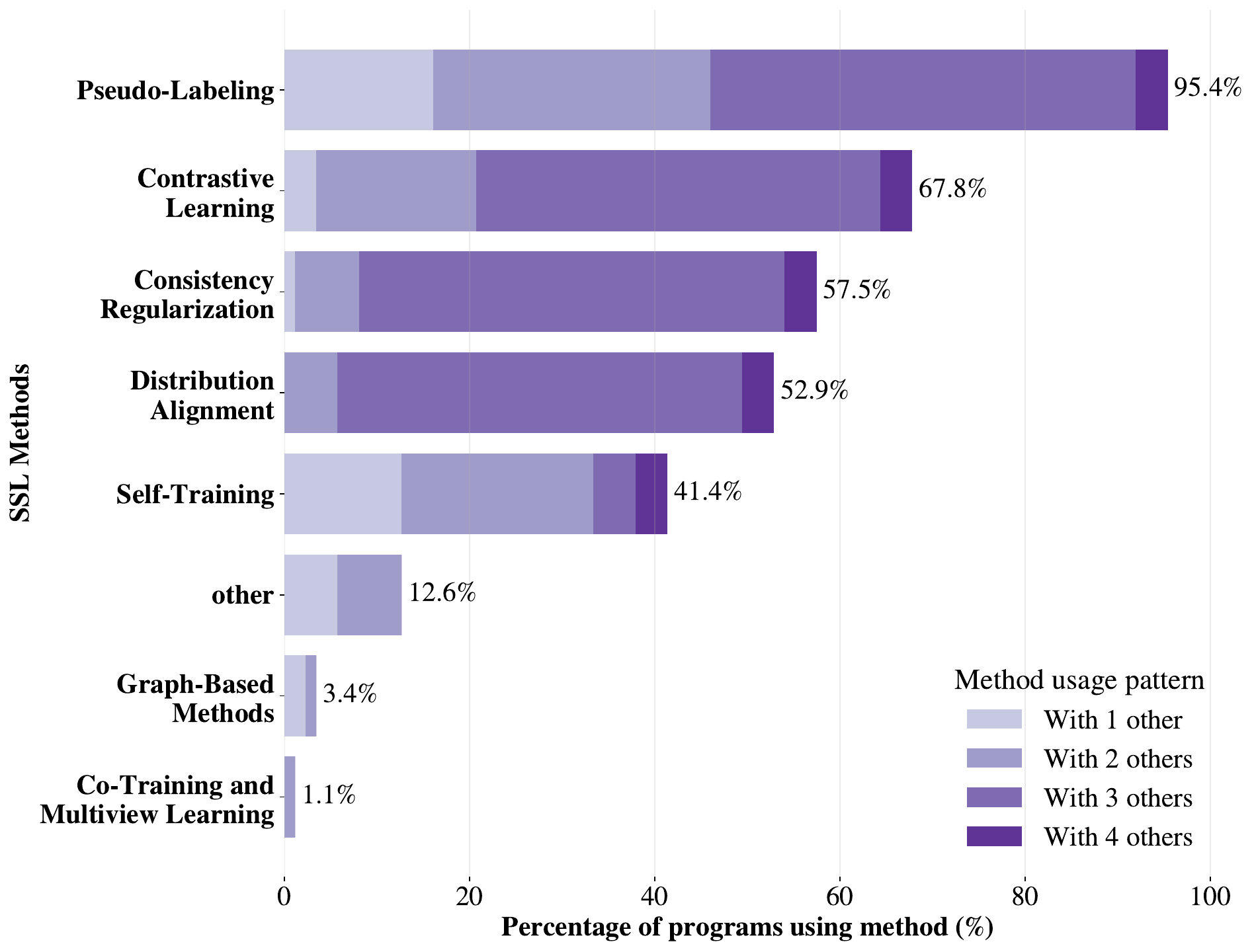}
        \caption{RESISC45}
        \label{fig:resisc_hlmd}
    \end{subfigure}

    \vspace{0.5em}

    \begin{subfigure}[t]{0.32\textwidth}
        \centering
        \includegraphics[width=\linewidth]{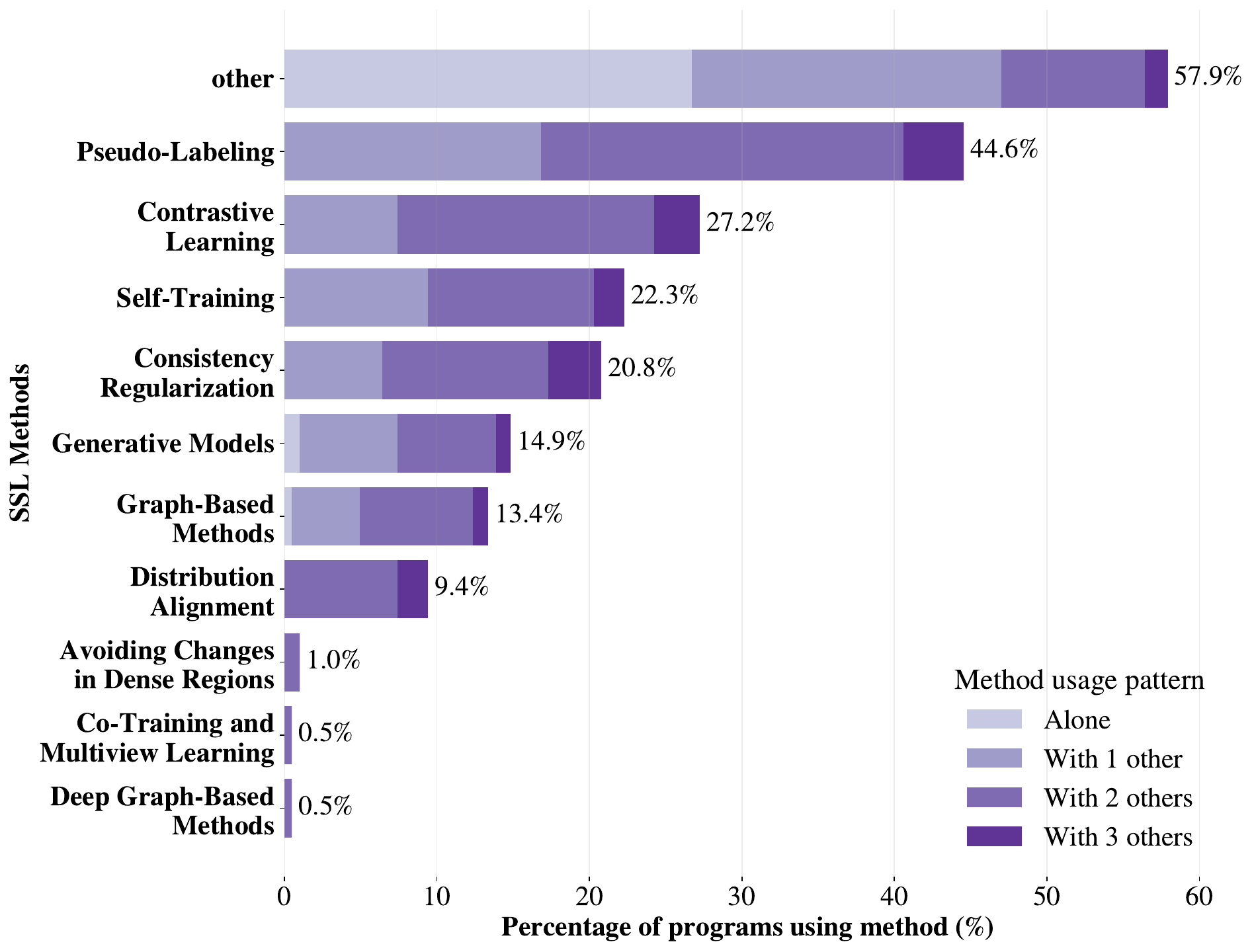}
        \caption{RETINO}
        \label{fig:retino_hlmd}
    \end{subfigure}
    \hfill
    \begin{subfigure}[t]{0.32\textwidth}
        \centering
        \includegraphics[width=\linewidth]{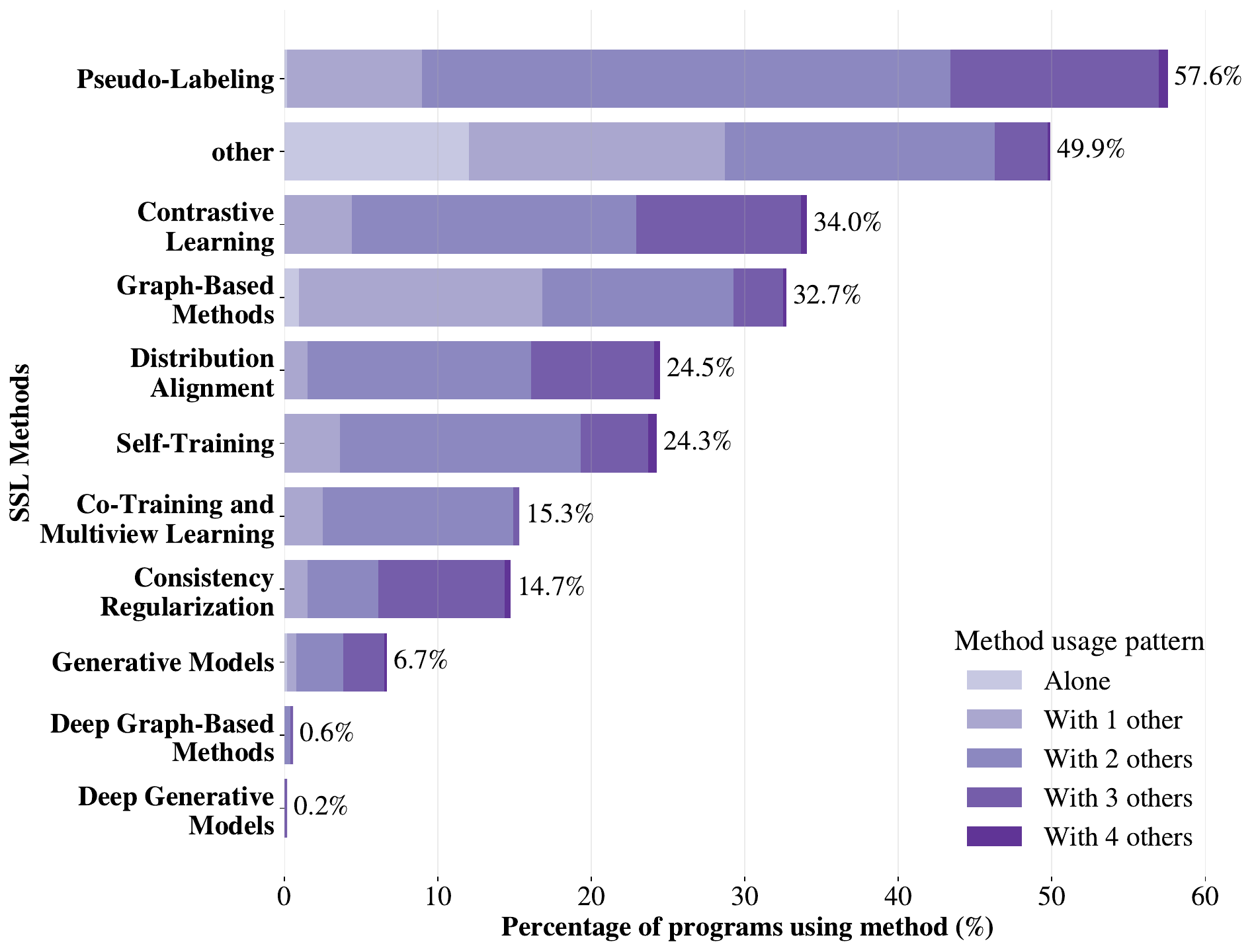}
        \caption{CLEVR}
        \label{fig:clevr_hlmd}
    \end{subfigure}
    \hfill
    \begin{subfigure}[t]{0.32\textwidth}
        \centering
        \includegraphics[width=\linewidth]{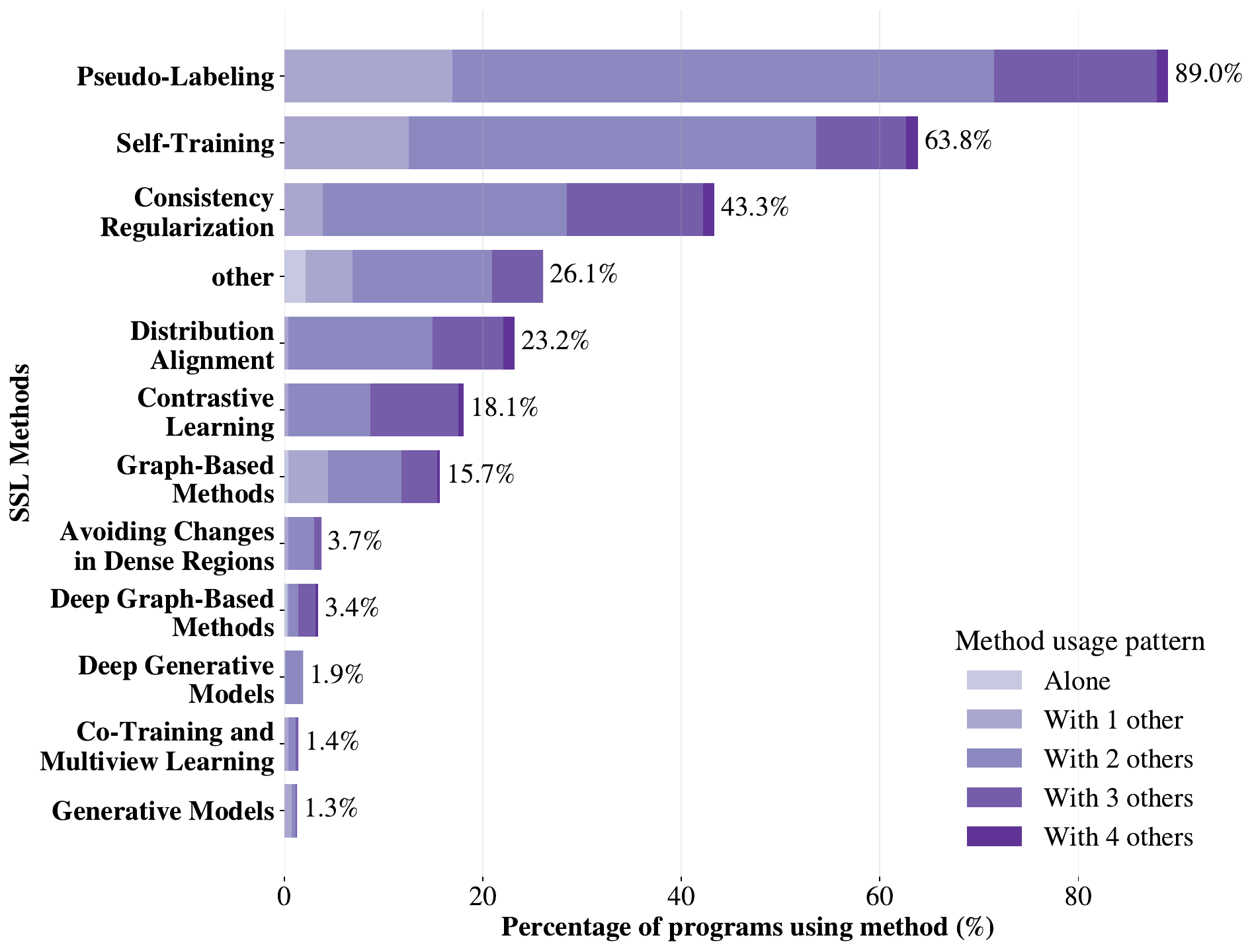}
        \caption{KITTI}
        \label{fig:kitti_hlmd}
    \end{subfigure}

    \caption{High-level SSL method distributions across datasets for greedy rollouts using the unsupervised metric as feedback.}
    \label{fig:all_hlmd}
\end{figure}

In this section, we show the high-level SSL method distributions for the remaining datasets under greedy rollouts using the unsupervised metric as feedback. The corresponding plot for DTD is included in the main paper, so we report the distributions for CLEVR, KITTI, RESISC45, SUN397, and Retinopathy in Figure~\ref{fig:all_hlmd}.

\section{Diversity Metric Details}
\label{appendix:diversity}

We measure the diversity of agent-generated programs following prior work
\cite{xuefei2025simpleagentsoutperformexperts}, adapting their API-based
Jaccard diversity metric to the high-level SSL method groups used in our
analysis.

For each program $i$, let $S_i$ denote the set of high-level SSL method groups
identified in that program by our LLM-as-judge pipeline. For two programs
$S_a$ and $S_b$, we compute their Jaccard dissimilarity as
\[
J_{\delta}(S_a,S_b)
=
1-\frac{|S_a\cap S_b|}{|S_a\cup S_b|}.
\]

Given $n$ generated programs, the final diversity score is the mean pairwise
Jaccard dissimilarity across all unique program pairs:
\[
\mathrm{Diversity}
=
\frac{1}{\binom{n}{2}}
\sum_{a<b} J_{\delta}(S_a,S_b).
\]

Higher values indicate that the agent explores programs composed of more
distinct combinations of SSL strategies, while lower values indicate that
the generated programs rely on more similar sets of SSL methods.

\section{Ecology Experiment Details}
\label{appendix:ecology_details}

\subsection{Datasets}
We select two ecological detection datasets to evaluate AgentSSL in real-world settings beyond classification. 

\paragraph{WildFin \cite{wildfin}} WildFin is an in-the-wild fish video behavior classification dataset that is derived from a real-world ecology pipeline. The detection artifact from there dataset was obtained by ecologists hand-labeling bounding boxes for three species of interest (requiring over 320 hours of expert effort, yielding 21,116 expert-labeled boxes). An ecologist implemented a program using YOLOv8, trained on this data, and used that detector for downstream tasks like cropping specific fish for behavior studies. We received their exact program and obtained our baseline results by executing them on our random subsets of their data. 

\paragraph{BuckTales \cite{naik2024bucktalesmultiuavdataset}} BuckTales is a UAV-based wildlife dataset focused on blackbuck monitoring and downstream multi-object tracking. The detection dataset contains approximately 24,000 bounding boxes across six object categories. While the full dataset includes additional classes such as dogs and birds for behavioral analysis, the original study focuses just on the blackbucks in the context of MOT and Re-ID. We obtained the program for their evaluation directly from the authors and re-ran their experiments on our subsets on \textbf{all classes}.

\subsection{Experiment Details}

For each dataset, we used three random seeds to randomly sample subsets of the full training sets (10\% of all images for WildFin and 20\% of all images for BuckTales). 

We execute the ecologist baseline program on all three of these random splits to produce the low-label domain expert baseline. We also execute the ecologist baseline programs on the entire training set to serve as the scientific success threshold, the threshold performance required to be actually used in the real-world scientific analysis pipeline.

We execute both AgentSL and AgentSSL using a greedy search policy and noisy validation feedback where 20\% of the data provided to the agent is set aside for validation. We run each of these agents once on each of the three random splits. 

\section{Agent SL vs. Agent SSL}
\label{appendix:SL_vs_SSL_app}

Although SSL research generally focuses on methods that effectively leverage unlabeled data, choosing not to use the unlabeled data and instead performing supervised learning (SL) is also a valid strategy. From this perspective, SL can be viewed as a subset of the broader SSL solution space. Indeed, prior work has shown that, in many settings, well-tuned SL can perform comparably to state-of-the-art SSL methods \cite{oliver2019realisticevaluationdeepsemisupervised}. To investigate this relationship in the context of AgentSSL, we conduct an ablation in which agents are not given access to unlabeled data. As shown in Table~\ref{appendix:agentsl_vs_agentssl}, under oracle selection, AgentSL performs comparably to AgentSSL, suggesting that the strongest supervised programs discovered by the agents can be as effective as those that explicitly leverage unlabeled data.

While AgentSSL is motivated in part by understanding how agents can leverage unlabeled data, our broader goal is to reduce the domain-specific engineering and annotation costs required to develop effective learning systems. If the best data-efficient solution for a particular task is simply a well-designed supervised learning pipeline, then this is a solution the agent should be able to discover. More broadly, we view the defining challenge of AgentSSL, and of the practical application of SSL more generally, as searching for effective programs in a regime with little or no validation data. Whether and how to leverage unlabeled data is therefore itself part of the search problem.

\input{tables/10_agentsl_vs_agentssl}

\section{Agent Prompts}
\label{appendix:agent_prompts}

\subsection{Operators}
\label{appendix:operators}
Our greedy search makes use of 4 operators: draft~\ref{fig:draft_prompt}, improve~\ref{fig:improve_prompt} and debug~\ref{fig:debug_prompt}, analyze~\ref{fig:program_analysis}. The prompts for the same are referenced here. 

\input{prompts/draft_operator}
\input{prompts/debug_operator}
\input{prompts/improve_operator}
\input{prompts/llm_analysis}

\subsection{Task Prompts}

All the Agent rollouts follow the same prompt structure for consistency. The prompts contain both task specific descriptions and dataset description. We provide the agent with a base class to then evolve during the rollout. The prompt also includes a description of this model interface as well as some information on how certain functions of this base class will be called during evalutation. 

\subsection{Image classification task}
Figure~\ref{fig:agentssl_prompt} and Figure~\ref{fig:agentsl_prompt} are two examples of image classification tasks for Agent SSL and Agent SL respectively where the given feedback is unsupervised metric. Figure~\ref{fig:agentssl_noisy_prompt} and Figure~\ref{fig:agentsl_noisy_prompt} are further examples for Agent SSL and Agent SL respectively when the feedback mechanism is noisy validation instead. 

\input{prompts/agent_ssl_prompt}
\input{prompts/agent_sl_prompt}
\input{prompts/agent_ssl_noisy_prompt}
\input{prompts/agent_sl_noisy_prompt}

\section{Example Agent Trajectories}
\label{appendix:agent_trajectories}

\begin{figure}[H]
    \centering
    \includegraphics[width=1\linewidth]{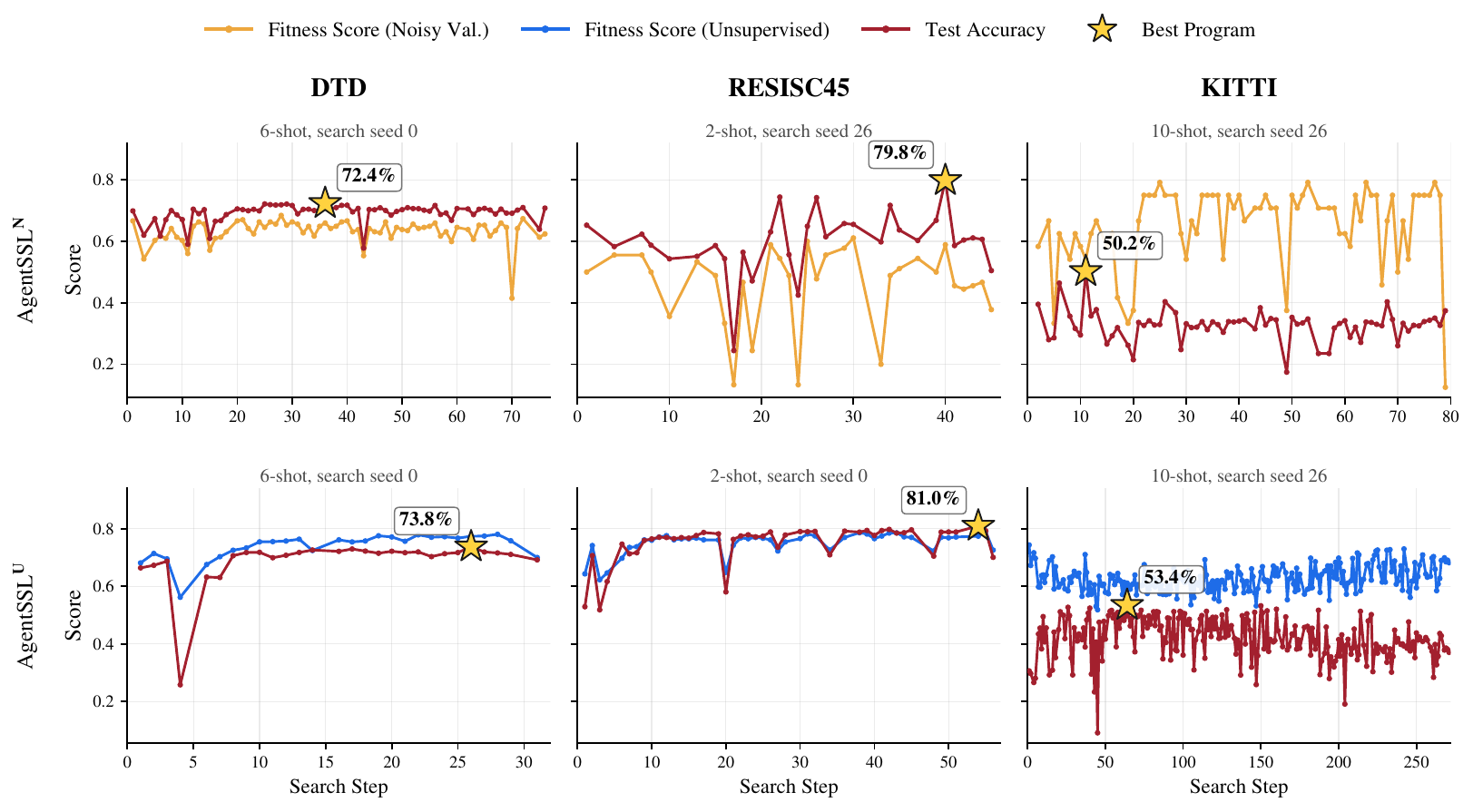}
    \caption{Comparison of rollout behavior under different feedback signals across datasets. 
Top row shows noisy validation feedback and bottom row shows unsupervised feedback. Fitness score denotes the scalar feedback metric provided to the agent during search.}
\label{fig:rollout_feedback_examples}
\end{figure}

\section{Interesting Programs}
\label{appendix:interesting_programs}
\subsection{Top-Performing Retinopathy Program}
\label{appendix:retino_best_program}
\input{figures/17_best_retino_program}

\section{Broader Societal Impact}
\label{appendix:societal_impact}
The potential negative social impacts of our method align with those typically associated with general LLM reasoning technologies. We emphasize the importance of adhering to the principles of fair and safe deployment in LLM systems.

%% file: tables/7_baseline_runtime.tex
\begin{table*}[ht]

\centering
\small
\setlength{\tabcolsep}{5pt}
\renewcommand{\arraystretch}{1.10}

\begin{tabular}{lrrrrrrr}
\toprule
\textbf{Algorithm}
& \textbf{DTD}
& \textbf{SUN397}
& \textbf{RESISC45}
& \textbf{RETINO}
& \textbf{CLEVR}
& \textbf{KITTI}
& \textbf{Total} \\
\midrule

Supervised & 56.7 & 272.1 & 74.1 & 194.2 & 165.3 & 30.3 & \textbf{792.6} \\
FixMatch   & 27.9 & 486.6 & 168.2 & 473.9 & 430.9 & 37.5 & \textbf{1,625.1} \\
SoftMatch  & 24.7 & 494.0 & 163.9 & 543.8 & 451.7 & 58.3 & \textbf{1,736.3} \\
FineSSL    & 23.9 & 800.4 & 154.8 & 1,293.1 & 400.8 & 42.7 & \textbf{2,715.7} \\
PET        & 5.7  & 158.0 & 43.7 & 91.2 & 83.4 & 9.6 & \textbf{391.6} \\
V-PET      & 7.6  & 139.5 & 42.0 & 98.4 & 70.8 & 9.2 & \textbf{367.4} \\

\midrule
\textbf{Baseline Total}
& \textbf{146.5}
& \textbf{2,350.6}
& \textbf{646.8}
& \textbf{2,694.5}
& \textbf{1,602.8}
& \textbf{187.5}
& \textbf{7,628.7} \\

\midrule
$\mathrm{AgentSSL}^{U}$
& 144.0 & 270.0 & 144.0 & 270.0 & 270.0 & 144.0 & \textbf{1242.0} \\

$\mathrm{AgentSSL}^{N}$
& 144.0 & 270.0 & 144.0 & 270.0 & 270.0 & 144.0 & \textbf{1242.0} \\
\midrule
\textbf{Agent Total}
& \textbf{288.0}
& \textbf{540.0}
& \textbf{288.0}
& \textbf{540.0}
& \textbf{540.0}
& \textbf{288.0}
& \textbf{2484.0} \\

\bottomrule
\end{tabular}

\caption{\small
Total \textbf{train} compute in GPU-hours. Baseline compute is aggregated across seeds,
shot settings, models, adaptation configurations, and learning rates.
Agent compute includes all trajectories using unsupervised feedback
($\mathrm{AgentSSL}^{U}$) and noisy-validation feedback
($\mathrm{AgentSSL}^{N}$).
}
\label{tab:baseline_runtime_totals}

\end{table*}

%% file: prompts/llm_as_judge_prompt.tex
\begin{promptbox}
\textbf{LLM-As-Judge Analysis Prompt}

\begin{lstlisting}
"""
You are an expert in data-efficient machine learning, semi-supervised learning, and program analysis.

Given a program, extract concise and accurate descriptors of:

1. Models used:
   Backbone architectures, pretrained models, model families, heads, or classifiers.
   Examples: CLIP, DINOv2

2. High-level method groups:
   Extract the PRIMARY semi-supervised learning group(s) used by the program.
   Use ONLY the compact method groups listed below. If none of the method groups are applicable, use "other".

3. Implementation details:
   Extract concrete mechanisms, design choices, losses, filters, calibration steps, feature processing steps,
   thresholds, scoring rules, optimization details, or architectural details used to implement the high-level method group(s).

4. Training usage:
   Add a one-word binary label on whether the program trains any model, classifier, adapter, head,
   student model, prototype module, or learnable parameter.

Return ONLY valid JSON in this format:

{
    "models": ["model1", "model2"],
    "high_level_methods": ["method_group1", "method_group2"],
    "implementation_details": ["detail1", "detail2"],
    "uses training": "yes" or "no",
    "summary": "A concise summary explaining why you chose the high level method groups and implementation details and how the program uses them. 1-5 sentences."
}

High-Level Method Groups:
Use these broad method groups whenever applicable.

'''
Generative Models (Old)
Graph-Based Methods (Old)
Avoiding Changes in Dense Regions
Distribution Alignment
Self-Training
Co-Training and Multiview Learning
Deep Generative Models
Consistency Regularization
Deep Graph-Based Methods
Pseudo-Labeling
Contrastive Learning
'''

Guidelines:
- Keep high_level_methods very coarse. Only use the method groups above that are applicable to the program. If none of the method groups are applicable, use "other".
- Use implementation_details to capture the specific algorithmic variant and other interesting details. Each high level method group contains the following details: 
    - Generative Models (Old):  Identifiability, Model Correctness, EM Local Maxima, Cluster-and-Label, Fisher kernel for discriminative learning
    - Graph-Based Methods (Old): KNN, Label Propagation, KNN Propagation, Mincut, Discrete Markov Random Fields: Boltzmann Machines, Gaussian Random Fields and Harmonic Functions, Tikhonov Regularization, Manifold Regularization, Laplacian Graph Kernels, Spectral Graph Transducer, Tree-Based Bayes
    - Avoiding Changes in Dense Regions: Transductive SVMs, Gaussian Processes, Information Regularization, Entropy Minimization
    - Self-Training.
    - Co-Training and Multiview Learning.
    - Deep Generative Methods: Semi-supervised GANs, Semi-supervised VAE, Semi-supervised Diffusion Models
    - Consistency Regularization: Ladder Network, PI-Model, Temporal Ensembling, Mean Teacher, VAT, Dual Student, UDA, etc. 
    - Deep Graph Based Methods: GCNN, GCN, GAT, GraphSAGE, etc.
    - Pseudo-Labeling: Confidence Thresholding, Hard Pseudo-Labels, Soft Pseudo-Labels, etc.
    - Contrastive Learning. 

- Include backbone or model family names when identifiable.
- Include classifier/head names in models when they are central to the method.
- Avoid adding many high-level groups unless the program truly combines multiple broad SSL paradigms.
- If the program uses a method outside the 5 groups and it is important, include a concise canonical name in high_level_methods.
- Do not include generic programming utilities, file handling, dataloading, logging, plotting, or evaluation code as methods.
- Do not include generic losses like cross-entropy unless they are modified or methodologically important.
- Do not include optimizers such as AdamW or SGD unless the optimizer choice is central to the method.
- Do not include preprocessing steps unless they materially affect the learning method.

Summary guidelines:
- Keep the summary concise, accurate, and method-focused.

Training label guidelines:
- "yes" if the program performs gradient-based training, fits a classifier, trains an adapter/head, updates a student model, performs linear probing, fits logistic regression/SVM/ridge classifiers, or learns parameters from data.
- "no" if the program only performs fixed feature extraction, nearest-neighbor lookup, prototype computation without learned parameters, deterministic graph propagation, or non-parametric inference.
"""
\end{lstlisting}
\label{fig:llm_as_judge_prompt}
\end{promptbox}

%% file: tables/10_agentsl_vs_agentssl.tex
\begin{table}[H]

\centering

\small

\captionsetup{width=0.78\linewidth}

\setlength{\tabcolsep}{4pt}

\renewcommand{\arraystretch}{1}

\begin{tabular}{l|cc|cc|cc}
\toprule
 & \multicolumn{2}{c}{\textbf{DTD}} 
 & \multicolumn{2}{c}{\textbf{RESISC45}} 
 & \multicolumn{2}{c}{\textbf{KITTI}} \\
\cmidrule(lr){2-3} 
\cmidrule(lr){4-5} 
\cmidrule(lr){6-7}

\textbf{Method} & 3 & 6 & 1 & 2 & 5 & 10 \\
\midrule

$\mathrm{AgentSL}_{avg}^{U}$   & 65.15 & 69.73 & 66.09 & 78.86 & 43.27 & 46.65 \\
$\mathrm{AgentSSL}_{avg}^{U}$  & 65.44 & 72.32 & 71.53 & 78.10 & 47.35 & 47.40 \\

$\mathrm{AgentSL}_{avg}^{N}$   & 65.53 & 72.55 & - & 73.71 & 44.16 & 40.23 \\
$\mathrm{AgentSSL}_{avg}^{N}$  & 65.57 & 71.45 & - & 76.06 & 42.71 & 45.85 \\

$\mathrm{AgentSL}_{max}^{U}$   & 65.83 & 72.23 & 74.83 & 79.56 & 37.26 & 40.50 \\
$\mathrm{AgentSSL}_{max}^{U}$  & 68.44 & 73.79 & 76.20 & 80.94 & 40.09 & 47.53 \\

$\mathrm{AgentSL}_{max}^{N}$   & 65.88 & 71.34 & - & 73.47 & 37.22 & 44.72 \\
$\mathrm{AgentSSL}_{max}^{N}$  & 64.71 & 72.15 & - & 76.93 & 39.19 & 42.84 \\

\bottomrule
\end{tabular}
\vspace{+4mm}
\caption{Comparison between AgentSL and AgentSSL across datasets and label settings. Superscripts indicate the feedback signal used during search: $U$ denotes unsupervised metrics and $N$ denotes noisy validation feedback. Subscripts indicate seed aggregation: $\mathrm{avg}$ reports average performance across seeds, while $\mathrm{max}$ reports the average best-discovered program from each seed.}

\label{appendix:agentsl_vs_agentssl}

\end{table}

%% file: prompts/draft_operator.tex
\begin{promptbox}
\textbf{Draft operator prompt}

\begin{lstlisting}
You are an expert in data-efficient machine learning and Python programming who is tasked with writing a data-efficient machine learning program to learn from a very small dataset with very few labels.

Your goal is to provide EXACTLY ONE IDEA AND ONE CODE IMPLEMENTATION of the idea, different from those previously explored, that leverages the available resources and is likely to lead to strong performance on the task.

Be specific about each step of the proposed approach, including data preprocessing, model architecture, training procedure, and prediction strategy.

You MUST PROVIDE a solution IDEA/PLAN in natural language and CODE in Python that DOES NOT INVOLVE any exploratory data analysis.

# TASK DESCRIPTION
{{task_desc}}

# PREVIOUSLY EXPLORED IDEAS
{{memory}}

# CONSTRAINTS
- Be aware of the running time of the solution; it should complete within {{execution_timeout}}.
- Prefer vectorized operations over Python loops when processing large datasets.
- Use torch.optim.AdamW instead of the deprecated AdamW from transformers.
- As much as possible, keep stdout clean.
- Print relevant training information such as loss, accuracy, etc.
- As this is the initial draft, prioritize speed in your solution.

# DATA
The training and validation data is already prepared and available in the read-only /data directory. You should not unzip any files.

# COMPUTE
You have access to a Python environment and the following packages installed: {{packages}}. DO NOT DOWNLOAD NEW PACKAGES.

Consider the previously explored ideas, and make sure the idea you propose considers a DIFFERENT ASPECT OF THE SOLUTION.

For example, you could try changing the model architecture. Brainstorm possible approaches and explain why they are likely to be effective and improve performance for the given task, data, and compute resources.

Remember: the first idea should be simple and easy to implement, while the last one should be more complex and sophisticated.

{% if draft_complexity == 'simple' %}
In this iteration, focus on PROPOSING A SIMPLE IDEA: one that can serve as a simple yet effective baseline for the task. For example, consider battle-tested methods or potentially pre-trained models known to work well for the task.
{% elif draft_complexity == 'normal' %}
In this iteration, focus on PROPOSING A MORE COMPLEX IDEA: one that can beat the previous baselines at the cost of some complexity and compute. For example, consider larger potentially pre-trained models, specialized feature engineering, basic ensembling, or hyperparameter optimization.
{% elif draft_complexity == 'complex' %}
In this iteration, focus on PROPOSING AN ADVANCED IDEA: one that can beat the previous baselines at the cost of some complexity and compute. For example, consider specialized potentially pre-trained models, advanced feature engineering or data augmentation strategies, advanced ensembling, or hyperparameter optimization.
{% endif %}

# RESPONSE FORMAT FOR IMPLEMENTATION

Provide a SINGLE Markdown code block for the implementation containing a SELF-CONTAINED Python script that implements the idea END-TO-END.

Start by making sure you understand the task, the data, the compute resources, and the idea. Then generate a detailed implementation plan that will guide the implementation step by step.

Reflect on the plan to ensure that the implementation is efficient and faithful to the idea, and that all requirements, including printing the evaluation score, are satisfied.

For large datasets, avoid for loops when possible and use efficient, fast data loading.

Format the proposed solution as follows:

# Idea to implement
<the proposed idea/plan>

```python
<the implementation of the proposed idea/plan>
\end{lstlisting}
\label{fig:draft_prompt}
\end{promptbox}

%% file: prompts/debug_operator.tex
\begin{promptbox}
\textbf{Debug operator prompt}

\begin{lstlisting}
  # Introduction:
  You are an expert in data-efficient machine learning and python programming who is tasked with
  fixing code bugs in a data-efficient machine learning program.
  Carefully review the previous debugging attempts, the buggy code and its terminal output in addition to the given task/data details, and available compute resources.
  You must not change the core idea or methodology of the solution, but only fix the bugs in the code.

  # Task Description:
  ````markdown
  {{task_desc}}
  ````

  {% if memory %}
  # Previous debugging attempts:
  ````markdown
  {{memory}}
  ````
  {% endif %}

  # Buggy Implementation:
  {{prev_buggy_code}}

  # Execution Output (Error):
  {{execution_output}}

  # Compute Environment:
  - Available Packages: {{packages}}
  - DO NOT DOWNLOAD NEW PACKAGES (You have access to pycocotools, torch, torchvision, ultralytics, etc.).

  # Instructions:
  - **Do NOT** alter the core method or underlying idea. Only correct existing bugs.
  - Outline your bug-fix plan clearly in 1-3 concise sentences.
  - Provide a single, complete Python code block wrapped in markdown (```python) that:
    - Implements the fix fully.
    - Is fully self-contained and executable as-is (The entire bug-free solution is given).

  - **Important Reminders:**
    - Absolutely do **NOT** skip any part of the code.
    - Make sure to print relevant training related information like the loss, accuracy, etc. 
    - Try to make the code run quickly and efficiently to get a quick proof of concept of your solution to the bug.

  Brainstorm about possible ways to fix the bug and WHY THEY ARE LIKELY TO FIX THE BUG for the given implementation. Additionally, if any other bugs further down the line are observed, please fix them as well.
  Generate a bug-fix plan that will structure and guide your step-by-step reasoning process. Reflect on it to make sure all the requirements are satisfied.

  Format the proposed bug-fix plan and code as follows:
  # Bug Fix Plan
  <bug-fix plan>
  
  ```python
  <the fixed python code>
  ```

\end{lstlisting}
\label{fig:debug_prompt}
\end{promptbox}

%% file: prompts/improve_operator.tex
\begin{promptbox}
\textbf{Improve operator prompt}

\begin{lstlisting}
# Introduction:
You are an expert in data-efficient machine learning and python programming who is tasked with
improving a data-efficient machine learning program to learn from a very small dataset with very few labels.
Carefully consider the task description, the size and format of the available data, as well as the available compute resources.
Your goal is to provide EXACTLY ONE IDEA AND ONE CODE IMPLEMENTATION of the idea, **different from those previously explored**, that improves upon an existing solution to the task.
Be specific about each step of the proposed improvement, including data processing and feature engineering and the modeling and optimization method
You MUST PROVIDE an improvement IDEA/PLAN in natural language and CODE in python that DOES NOT INVOLVE any exploratory data analysis.   

# TASK DESCRIPTION
````
{{task_desc}}
````

# PREVIOUS SOLUTION:
## Code:
{{prev_code}}
## Terminal Output:
{{prev_terminal_output}}

# PREVIOUSLY EXPLORED IMPROVEMENT IDEAS
````markdown
{{memory}}
````

**CONSTRAINTS**:
- Be aware of the running time of the solution, it should complete within {{execution_timeout}}
- Prefer vectorized operations over Python loops when processing large datasets.  
- Use `torch.optim.AdamW` (the recommended optimizer) instead of the deprecated `AdamW` from `transformers`.  
- Make sure to print relevant training related information like the loss, accuracy, etc. 
- As much as possible, keep the stdout clean.

**COMPUTE**: You have access to a Python environemnt with the following packages installed: {{packages}}. DO NOT DOWNLOAD NEW PACKAGES.

Consider the previously explored ideas, and make sure the improvement idea you propose considers a DIFFERENT IMPROVEMENT OF THE SOLUTION. 
Brainstorm about possible improvements and WHY THEY ARE LIKELY TO BE EFFECTIVE AND INCREASE THE PERFORMANCE for the given task, and the available data and compute resources.
{% if improve_complexity == 'simple' %}
In this iteration, suggest a *minimal, low-risk* tweak that keeps the current solution's core intact - no architecture overhauls or fundamental methodology changes.  
Try to keep the training time short and efficient to iterate quickly.
Check the MEMORY section first and avoid duplicating earlier ideas. 
{% elif improve_complexity == 'normal' %}
In this iteration, propose a *moderate upgrade* that builds on the baseline without deviating dramatically. Can increase the training time a bit more than before. 
Check the MEMORY section first and avoid duplicating earlier ideas.
{% elif improve_complexity == 'complex' %}
In this iteration, recommend a *substantial extension* that pushes the method's boundaries while preserving its core logic.  
Check the MEMORY section first and avoid duplicating earlier ideas.
{% endif %}

**RESPONSE FORMAT FOR IMPLEMENTATION**: 
Provide a **SINGLE** Markdown code block (wrapped in ```) containing a **SELF-CONTAINED** Python script that:
1. Implements the idea **END-TO-END**

Start by making sure you understand the task, the data and compute resources and the idea. Then generate a detailed implementation plan that will structure and guide you step-by-step through the implementation process. Make sure to reflect on the plan to ensure that the implementation is efficient and faithful to the idea, and that all the requirements (e.g., the evaluation score is printed, the submission file follows the correct format and is saved in the correct location, etc.) are satisfied.
Format the proposed solution as follows:

# Improvement Idea to implement
<the proposed improvement idea/plan>

```python
<the implementation of the proposed improvement>
```
\end{lstlisting}
\label{fig:improve_prompt}
\end{promptbox}

%% file: prompts/llm_analysis.tex
\begin{promptbox}
\textbf{Analysis Operator}

\begin{lstlisting}
      # Introduction:
      You are an expert in data-efficient machine learning and python programming who is tasked with
      writing a data-efficient machine learning program to learn from a very small dataset with very few labels. 
      You have written code to solve this task and now need to evaluate the output of the code execution.
      You should determine if there were any bugs as well as report the empirical findings. Please keep this 
      evaluation extremely short and concise (1-3 relevant bullet points).

      # Task Description:
      {{task_desc}}

      # Implementation:
      {{code}}

      # Execution output:
      {{execution_output}}
"""
\end{lstlisting}
\label{fig:program_analysis}
\end{promptbox}

%% file: prompts/agent_ssl_prompt.tex
\begin{center}
\captionsetup{type=figure}

\begin{promptbox}

\textbf{Example Agent SSL prompt}

\begin{lstlisting}

# Objective
You are an AI agent tasked with implementing a classification program for an image dataset. Your goal is to train (or otherwise derive) a classifier using the provided training dataset and produce predicted labels for every image in the validation dataset.

You have access to a very small training set, a validation set and a set of unlabelled datapoints. Your output will be evaluated using a standardized evaluation script which will return a set of unsupervised proxy metrics (ami, ari, v_measure, fmi, silhouette) as the feedback, but note that the priority is test set performance. At the end of the evolution, the final program will be evaluated on the validation set.

# Dataset specification
A texture recognition dataset consisting of real-world images annotated with human-interpretable texture attributes. It contains 47 categories (e.g., striped, dotted, fibrous, cracked). The images exhibit significant variation in scale, illumination, viewpoint, and background clutter. Your training dataset contains 3 images per class for a total of just 141 images. Your unlabeled dataset has 1739 datapoints.

# Model Interface
You MUST use the provided model builder.

1. Choose backbone:
    net_name in {
        "timm/vit_base_patch16_clip_224.openai",
        "timm/vit_base_patch14_reg4_dinov2.lvd142m"
    }
2. Create PEFT config:
    peft_config = get_peft_config_fn({...})

PEFT options:
Using Lora:
    {"method_name": "lora_1", "lora_bottleneck": 4}
Using Adaptformer:
    {"method_name": "adaptformer", "ft_mlp_module": "adapter", "ft_mlp_mode": "parallel", "ft_mlp_ln": "before", "adapter_init": "lora_kaiming", "adapter_bottleneck": 4, "adapter_scaler": 0.1}
Frozen Backbone:
    {"freeze_backbone": True}
3. Set vit_config:
    vit_config = {"drop_path_rate": 0.2}
4. Build model class:
    net_builder = net_builder_fn( net_name, peft_config, vit_config)

IMPORTANT:
net_builder is a CLASS, not an instance.

5. Instantiate:
    self.model = net_builder( num_classes=num_classes, pretrained=True, pretrained_path="").to(device)

# Model Behavior

out = model(x) returns:
    out["feat"]   : pooled feature vector
    out["logits"] : classification logits

IMPORTANT:
Use:
    trainable = [
        p for p in self.model.parameters()
        if p.requires_grad
    ]
to identify optimizable parameters.

# Required Program Interface
class ClassificationAgent:

    def __init__( self, net_builder_fn, get_peft_config_fn, num_classes):
        self.transform = pass
        self.model = pass

    def fit(self, img_dir, train_ann_file, unlabel_ann_file):
        pass

    def predict(self, image_batch):
        pass

# Usage
program = load_program(program_path)
agent = program.ClassificationAgent( net_builder_fn=get_net_builder, get_peft_config=get_peft_config, num_classes=num_classes)
agent.fit( train_img_dir, train_ann_file, unlabel_ann_file)
image = agent.transform(raw_image_tensor)
logits, feat, predictions = agent.predict(image)

\end{lstlisting}
\end{promptbox}
\caption{ Prompt provided for an example AgentSSL rollout for dataset DTD. The prompt includes task specific description including the feedback metric (in objective, model interface and usage section) and dataset specific description (in the dataset specification section). The above example is for dataset DTD for the 3-shot partition. 
}
\label{fig:agentssl_prompt}
\end{center}

%% file: prompts/agent_sl_prompt.tex
\begin{center}
\captionsetup{type=figure}
\begin{promptbox}

\textbf{Example Agent SL prompt}

\begin{lstlisting}

# Objective
You are an AI agent tasked with implementing a **classification program** for an image dataset. Your goal is to train (or otherwise derive) a classifier using the provided training dataset and produce **predicted labels for every image in the validation dataset**.

You have access to a very small training set, a validation set. Your output will be evaluated using a standardized evaluation script which will return a set of unsupervised proxy metrics (ami, ari, v_measure, fmi, silhouette) as the feedback, but note that the priority is test set performance. At the end of the evolution, the final program will be evaluated on the validation set.

# Dataset specification
A texture recognition dataset consisting of real-world images annotated with human-interpretable texture attributes. It contains 47 categories (e.g., striped, dotted, fibrous, cracked). The images exhibit significant variation in scale, illumination, viewpoint, and background clutter. Your training dataset contains 3 images per class for a total of just 141 images. Your unlabeled dataset has 1739 datapoints.

# Model Interface
You MUST use the provided model builder.

1. Choose backbone:
    net_name \in {
        "timm/vit_base_patch16_clip_224.openai",
        "timm/vit_base_patch14_reg4_dinov2.lvd142m"
    }
2. Create PEFT config:
    peft_config = get_peft_config_fn({...})

PEFT options:
Using Lora:
    {"method_name": "lora_1", "lora_bottleneck": 4}
Using Adaptformer:
    {"method_name": "adaptformer", "ft_mlp_module": "adapter", "ft_mlp_mode": "parallel", "ft_mlp_ln": "before", "adapter_init": "lora_kaiming", "adapter_bottleneck": 4, "adapter_scaler": 0.1}
Frozen Backbone:
    {"freeze_backbone": True}
3. Set vit_config:
    vit_config = {"drop_path_rate": 0.2}
4. Build model class:
    net_builder = net_builder_fn( net_name, peft_config, vit_config)

IMPORTANT:
net_builder is a CLASS, not an instance.

5. Instantiate:
    self.model = net_builder( num_classes=num_classes, pretrained=True, pretrained_path="").to(device)

# Model Behavior

out = model(x) returns:
    out["feat"]   : pooled feature vector
    out["logits"] : classification logits

IMPORTANT:
Use:
    trainable = [
        p for p in self.model.parameters()
        if p.requires_grad
    ]
to identify optimizable parameters.

# Required Program Interface
class ClassificationAgent:

    def __init__( self, net_builder_fn, get_peft_config_fn, num_classes):
        self.transform = pass
        self.model = pass

    def fit(self, img_dir, train_ann_file):
        pass

    def predict(self, image_batch):
        pass

# Usage
program = load_program(program_path)
agent = program.ClassificationAgent( net_builder_fn=get_net_builder, get_peft_config=get_peft_config, num_classes=num_classes)
agent.fit( train_img_dir, train_ann_file)
image = agent.transform(raw_image_tensor)
logits, feat, predictions = agent.predict(image)

\end{lstlisting}
\end{promptbox}
\caption{ Prompt provided for an example AgentSL rollout for dataset DTD. The prompt includes task specific description including the feedback metric (in objective, model interface and usage section) and dataset specific description (in the dataset specification section). The above example is for dataset DTD for the 3-shot partition. 
}
\label{fig:agentsl_prompt}
\end{center}

%% file: prompts/agent_ssl_noisy_prompt.tex
\begin{center}
\captionsetup{type=figure}

\begin{promptbox}

\textbf{Example Agent SSL prompt}

\begin{lstlisting}

# Objective
You are an AI agent tasked with implementing a classification program for an image dataset. Your goal is to train (or otherwise derive) a classifier using the provided training dataset and produce predicted labels for every image in the validation dataset.

You have access to a very small training set, a validation set and a set of unlabelled datapoints. Your output will be evaluated using a standardized evaluation script which will return a limited proxy validation metric as the feedback, but note that the priority is test set performance. At the end of the evolution, the final program will be evaluated on the full validation set.

# Dataset specification
A texture recognition dataset consisting of real-world images annotated with human-interpretable texture attributes. It contains 47 categories (e.g., striped, dotted, fibrous, cracked). The images exhibit significant variation in scale, illumination, viewpoint, and background clutter. Your training dataset contains 3 images per class for a total of just 141 images. Your unlabeled dataset has 1739 datapoints.

# Model Interface
You MUST use the provided model builder.

1. Choose backbone:
    net_name \in {
        "timm/vit_base_patch16_clip_224.openai",
        "timm/vit_base_patch14_reg4_dinov2.lvd142m"
    }
2. Create PEFT config:
    peft_config = get_peft_config_fn({...})

PEFT options:
Using Lora:
    {"method_name": "lora_1", "lora_bottleneck": 4}
Using Adaptformer:
    {"method_name": "adaptformer", "ft_mlp_module": "adapter", "ft_mlp_mode": "parallel", "ft_mlp_ln": "before", "adapter_init": "lora_kaiming", "adapter_bottleneck": 4, "adapter_scaler": 0.1}
Frozen Backbone:
    {"freeze_backbone": True}
3. Set vit_config:
    vit_config = {"drop_path_rate": 0.2}
4. Build model class:
    net_builder = net_builder_fn( net_name, peft_config, vit_config)

IMPORTANT:
net_builder is a CLASS, not an instance.

5. Instantiate:
    self.model = net_builder( num_classes=num_classes, pretrained=True, pretrained_path="").to(device)

# Model Behavior

out = model(x) returns:
    out["feat"]   : pooled feature vector
    out["logits"] : classification logits

IMPORTANT:
Use:
    trainable = [
        p for p in self.model.parameters()
        if p.requires_grad
    ]
to identify optimizable parameters.

# Required Program Interface
class ClassificationAgent:

    def __init__( self, net_builder_fn, get_peft_config_fn, num_classes):
        self.transform = pass
        self.model = pass

    def fit(self, img_dir, train_ann_file, unlabel_ann_file):
        pass

    def predict(self, image_batch):
        pass

# Usage
program = load_program(program_path)
agent = program.ClassificationAgent( net_builder_fn=get_net_builder, get_peft_config=get_peft_config, num_classes=num_classes)
agent.fit( train_img_dir, train_ann_file, unlabel_ann_file)
image = agent.transform(raw_image_tensor)
logits, feat, predictions = agent.predict(image)

\end{lstlisting}
\end{promptbox}
\caption{ Prompt provided for an example AgentSSL rollout for dataset DTD. The prompt includes task specific description including the feedback metric (in objective, model interface and usage section) and dataset specific description (in the dataset specification section). The above example is for dataset DTD for the 3-shot partition where the feedback given is noisy validation signal. 
}
\label{fig:agentssl_noisy_prompt}
\end{center}

%% file: prompts/agent_sl_noisy_prompt.tex
\begin{center}
\captionsetup{type=figure}

\begin{promptbox}

\textbf{Example Agent SL prompt}

\begin{lstlisting}

# Objective
You are an AI agent tasked with implementing a **classification program** for an image dataset. Your goal is to train (or otherwise derive) a classifier using the provided training dataset and produce **predicted labels for every image in the validation dataset**.

You have access to a very small training set, a validation set and a set of unlabelled datapoints. Your output will be evaluated using a standardized evaluation script which will return a limited proxy validation metric as the feedback, but note that the priority is test set performance. At the end of the evolution, the final program will be evaluated on the full validation set.

# Dataset specification
A texture recognition dataset consisting of real-world images annotated with human-interpretable texture attributes. It contains 47 categories (e.g., striped, dotted, fibrous, cracked). The images exhibit significant variation in scale, illumination, viewpoint, and background clutter. Your training dataset contains 3 images per class for a total of just 141 images. Your unlabeled dataset has 1739 datapoints.

# Model Interface
You MUST use the provided model builder.

1. Choose backbone:
    net_name \in {
        "timm/vit_base_patch16_clip_224.openai",
        "timm/vit_base_patch14_reg4_dinov2.lvd142m"
    }
2. Create PEFT config:
    peft_config = get_peft_config_fn({...})

PEFT options:
Using Lora:
    {"method_name": "lora_1", "lora_bottleneck": 4}
Using Adaptformer:
    {"method_name": "adaptformer", "ft_mlp_module": "adapter", "ft_mlp_mode": "parallel", "ft_mlp_ln": "before", "adapter_init": "lora_kaiming", "adapter_bottleneck": 4, "adapter_scaler": 0.1}
Frozen Backbone:
    {"freeze_backbone": True}
3. Set vit_config:
    vit_config = {"drop_path_rate": 0.2}
4. Build model class:
    net_builder = net_builder_fn( net_name, peft_config, vit_config)

IMPORTANT:
net_builder is a CLASS, not an instance.

5. Instantiate:
    self.model = net_builder( num_classes=num_classes, pretrained=True, pretrained_path="").to(device)

# Model Behavior

out = model(x) returns:
    out["feat"]   : pooled feature vector
    out["logits"] : classification logits

IMPORTANT:
Use:
    trainable = [
        p for p in self.model.parameters()
        if p.requires_grad
    ]
to identify optimizable parameters.

# Required Program Interface
class ClassificationAgent:

    def __init__( self, net_builder_fn, get_peft_config_fn, num_classes):
        self.transform = pass
        self.model = pass

    def fit(self, img_dir, train_ann_file):
        pass

    def predict(self, image_batch):
        pass

# Usage
program = load_program(program_path)
agent = program.ClassificationAgent( net_builder_fn=get_net_builder, get_peft_config=get_peft_config, num_classes=num_classes)
agent.fit( train_img_dir, train_ann_file)
image = agent.transform(raw_image_tensor)
logits, feat, predictions = agent.predict(image)

\end{lstlisting}
\end{promptbox}
\caption{ Prompt provided for an example AgentSL rollout for dataset DTD. The prompt includes task specific description including the feedback metric (in objective, model interface and usage section) and dataset specific description (in the dataset specification section). The above example is for dataset DTD for the 3-shot partition where the feedback given is a noisy validation signal. 
}
\label{fig:agentsl_noisy_prompt}
\end{center}

%% file: figures/17_best_retino_program.tex
\begin{promptbox}
\textbf{Best Retino Program}

\begin{lstlisting}
from __future__ import annotations
import json
import os
from typing import List, Tuple

import torch
import torch.nn.functional as F
from torch.utils.data import Dataset, DataLoader
import torchvision.transforms as T
from PIL import Image

def _read_json(path: str):
    with open(path, "r") as f:
        return json.load(f)

def _coco_label_mapping(train_ann_file: str):
    coco = _read_json(train_ann_file)
    cat_ids = sorted(c["id"] for c in coco["categories"])
    cat_id_to_idx = {cid: i for i, cid in enumerate(cat_ids)}
    idx_to_cat_id = {i: cid for cid, i in cat_id_to_idx.items()}
    return cat_id_to_idx, idx_to_cat_id, len(cat_ids)

def _labeled_records(img_dir: str, train_ann_file: str, cat_id_to_idx: dict):
    coco = _read_json(train_ann_file)
    id_to_name = {img["id"]: img["file_name"] for img in coco["images"]}
    out = []
    for ann in coco["annotations"]:
        fn = id_to_name.get(ann["image_id"])
        if fn is None:
            continue
        path = os.path.join(img_dir, fn)
        if not os.path.isfile(path):
            continue
        cid = ann["category_id"]
        if cid not in cat_id_to_idx:
            continue
        out.append((path, cat_id_to_idx[cid]))
    return out

def _unlabeled_records(img_dir: str, unlabel_ann_file: str):
    coco = _read_json(unlabel_ann_file)
    out = []
    for img in coco["images"]:
        fn = img["file_name"]
        path = os.path.join(img_dir, fn)
        if os.path.isfile(path):
            out.append(path)
    return out

class _ImageOnlyDataset(Dataset):
    def __init__(self, paths, transform):
        self.paths = paths
        self.transform = transform

    def __len__(self):
        return len(self.paths)

    def __getitem__(self, idx):
        img = Image.open(self.paths[idx]).convert("RGB")
        return self.transform(img)

class _GraphConfig:
    def __init__(
        self,
        k_ul: int = 24,
        k_uu: int = 24,
        temperature: float = 0.07,
        alpha: float = 0.85,
        iters: int = 25,
        chunk_q: int = 1024,
    ):
        self.k_ul = k_ul
        self.k_uu = k_uu
        self.temperature = temperature
        self.alpha = alpha
        self.iters = iters
        self.chunk_q = chunk_q

class ClassificationAgent:
    NET_NAME = "timm/vit_base_patch14_reg4_dinov2.lvd142m"

    def __init__(self, net_builder_fn, get_peft_config_fn, num_classes):
        self.net_builder_fn = net_builder_fn
        self.get_peft_config_fn = get_peft_config_fn
        self.num_classes = num_classes
        self.device = torch.device("cuda" if torch.cuda.is_available() else "cpu")

        self.img_size = 224
        self.transform = self._eval_transform(self.img_size)
        self.idx_to_cat_id = {i: i for i in range(num_classes)}

        peft_config = self.get_peft_config_fn({"freeze_backbone": True})
        vit_config = {"drop_path_rate": 0.0}
        net_builder = self.net_builder_fn(self.NET_NAME, peft_config, vit_config)
        self.model = net_builder(
            num_classes=num_classes, pretrained=True, pretrained_path=""
        ).to(self.device)
        self.model.eval()

        self.graph_cfg = _GraphConfig()

        self.train_feat = None
        self.train_labels = None
        self.train_prop = None
        self.class_prototypes = None
        self.residual_quantiles = None

        self.k_query = 32
        self.k_proto = 5
        self.tta_shift = 8

        self.fuse_w_knn_hard = 1.15
        self.fuse_w_knn_prop = 0.85
        self.fuse_w_proto = 0.55
        self.fuse_w_ord = 0.30
        self.score_temp = 0.07
        self.proto_temp = 0.09
        self.conformal_q = 0.80

    def _eval_transform(self, img_size=224):
        resize = int(img_size * 256 / 224)
        return T.Compose(
            [
                T.Resize(resize, interpolation=T.InterpolationMode.BICUBIC),
                T.CenterCrop(img_size),
                T.ToTensor(),
                T.Normalize(
                    mean=[0.485, 0.456, 0.406],
                    std=[0.229, 0.224, 0.225],
                ),
            ]
        )

    @torch.no_grad()
    def _extract_features(self, paths, batch_size=64):
        if len(paths) == 0:
            return torch.empty((0, 768), dtype=torch.float32)

        ds = _ImageOnlyDataset(paths, self.transform)
        loader = DataLoader(
            ds,
            batch_size=min(batch_size, max(1, len(ds))),
            shuffle=False,
            num_workers=min(8, os.cpu_count() or 0),
            pin_memory=self.device.type == "cuda",
            drop_last=False,
        )

        feats = []
        for x in loader:
            x = x.to(self.device, non_blocking=True)
            out = self.model(x)
            feats.append(out["feat"].detach().cpu())
        if len(feats) == 0:
            return torch.empty((0, 768), dtype=torch.float32)
        return torch.cat(feats, dim=0)

    def _build_sparse_neighbors(self, xu: torch.Tensor, xl: torch.Tensor):
        cfg = self.graph_cfg
        device = self.device
        xu = xu.to(device)
        xl = xl.to(device)

        Nu = xu.shape[0]
        Nl = xl.shape[0]
        if Nu == 0 or Nl == 0:
            raise RuntimeError(
                "Neighbor graph requires both labeled and unlabeled features."
            )

        k_ul = min(cfg.k_ul, Nl)
        k_uu = min(cfg.k_uu + 1, Nu)

        ul_idx_parts, ul_w_parts = [], []
        uu_idx_parts, uu_w_parts = [], []

        for start in range(0, Nu, cfg.chunk_q):
            end = min(start + cfg.chunk_q, Nu)
            q = xu[start:end]

            sim_ul = q @ xl.T
            vals_ul, idx_ul = torch.topk(sim_ul, k=max(1, k_ul), dim=1)
            w_ul = torch.softmax(vals_ul / cfg.temperature, dim=1)
            ul_idx_parts.append(idx_ul.cpu())
            ul_w_parts.append(w_ul.cpu())

            if Nu == 1:
                idx_uu = torch.zeros((end - start, 1), dtype=torch.long, device=device)
                vals_uu = torch.zeros(
                    (end - start, 1), dtype=torch.float32, device=device
                )
            else:
                sim_uu = q @ xu.T
                vals_uu, idx_uu = torch.topk(sim_uu, k=max(2, k_uu), dim=1)
                rows = torch.arange(start, end, device=device).unsqueeze(1)
                keep = idx_uu != rows
                idx_uu = idx_uu[keep].view(end - start, -1)
                vals_uu = vals_uu[keep].view(end - start, -1)
                if idx_uu.shape[1] == 0:
                    idx_uu = torch.zeros(
                        (end - start, 1), dtype=torch.long, device=device
                    )
                    vals_uu = torch.zeros(
                        (end - start, 1), dtype=torch.float32, device=device
                    )
                else:
                    idx_uu = idx_uu[:, : min(cfg.k_uu, idx_uu.shape[1])]
                    vals_uu = vals_uu[:, : min(cfg.k_uu, vals_uu.shape[1])]

            w_uu = torch.softmax(vals_uu / cfg.temperature, dim=1)
            uu_idx_parts.append(idx_uu.cpu())
            uu_w_parts.append(w_uu.cpu())

        ul_idx = torch.cat(ul_idx_parts, dim=0)
        ul_w = torch.cat(ul_w_parts, dim=0)
        uu_idx = torch.cat(uu_idx_parts, dim=0)
        uu_w = torch.cat(uu_w_parts, dim=0)
        return ul_idx, ul_w, uu_idx, uu_w

    def _label_propagation(self, xl: torch.Tensor, yl: torch.Tensor, xu: torch.Tensor):
        num_classes = self.num_classes
        cfg = self.graph_cfg

        xl = F.normalize(xl.float(), dim=1)
        xu = F.normalize(xu.float(), dim=1)

        y_onehot = F.one_hot(yl, num_classes=num_classes).float()
        ul_idx, ul_w, uu_idx, uu_w = self._build_sparse_neighbors(xu, xl)

        yu = torch.full(
            (xu.shape[0], num_classes), 1.0 / num_classes, dtype=torch.float32
        )

        for _ in range(cfg.iters):
            msg_l = torch.sum(ul_w.unsqueeze(-1) * y_onehot[ul_idx], dim=1)
            msg_u = torch.sum(uu_w.unsqueeze(-1) * yu[uu_idx], dim=1)
            yu = (1.0 - cfg.alpha) * msg_l + cfg.alpha * msg_u
            yu = yu / yu.sum(dim=1, keepdim=True).clamp_min(1e-8)

        return y_onehot, yu

    def _compute_train_propagated_probs(
        self, xl: torch.Tensor, yl: torch.Tensor, xu: torch.Tensor, yu: torch.Tensor
    ):
        xl_n = F.normalize(xl.float(), dim=1)
        xu_n = F.normalize(xu.float(), dim=1)
        y_onehot = F.one_hot(yl, num_classes=self.num_classes).float()

        sims_ll = xl_n @ xl_n.T
        sims_lu = xl_n @ xu_n.T

        if xl.shape[0] <= 1:
            msg_ll = y_onehot.clone()
        else:
            k_ll = min(16, xl.shape[0] - 1)
            vals_ll, idx_ll = torch.topk(
                sims_ll, k=min(k_ll + 1, sims_ll.shape[1]), dim=1
            )
            rows = torch.arange(xl.shape[0]).unsqueeze(1)
            keep = idx_ll != rows
            idx_ll = idx_ll[keep].view(xl.shape[0], -1)[:, :k_ll]
            vals_ll = vals_ll[keep].view(xl.shape[0], -1)[:, :k_ll]
            w_ll = torch.softmax(vals_ll / 0.07, dim=1)
            msg_ll = torch.sum(w_ll.unsqueeze(-1) * y_onehot[idx_ll], dim=1)

        if xu.shape[0] == 0:
            msg_lu = torch.full_like(msg_ll, 1.0 / self.num_classes)
        else:
            k_lu = min(32, xu.shape[0])
            vals_lu, idx_lu = torch.topk(sims_lu, k=k_lu, dim=1)
            w_lu = torch.softmax(vals_lu / 0.07, dim=1)
            msg_lu = torch.sum(w_lu.unsqueeze(-1) * yu[idx_lu], dim=1)

        prop = 0.6 * msg_ll + 0.4 * msg_lu
        prop = prop / prop.sum(dim=1, keepdim=True).clamp_min(1e-8)
        return prop

    def _build_prototypes(self, xl: torch.Tensor, yl: torch.Tensor):
        protos = []
        for c in range(self.num_classes):
            mask = yl == c
            if mask.any():
                proto = xl[mask].mean(dim=0)
            else:
                proto = xl.mean(dim=0)
            protos.append(F.normalize(proto, dim=0))
        return torch.stack(protos, dim=0)

    def _fused_local_logits(
        self,
        query_feat: torch.Tensor,
        leave_one_out_index: int | None = None,
    ):
        feat = F.normalize(query_feat.float(), dim=1).cpu()
        train_feat = self.train_feat
        train_labels = self.train_labels
        train_prop = self.train_prop
        prototypes = self.class_prototypes
        num_classes = self.num_classes

        sim = feat @ train_feat.T

        if leave_one_out_index is not None and feat.shape[0] == 1:
            sim[:, leave_one_out_index] = -1e4

        k = min(
            self.k_query,
            train_feat.shape[0] - (1 if leave_one_out_index is not None else 0),
        )
        k = max(1, k)

        vals, idx = torch.topk(sim, k=k, dim=1)
        w = torch.softmax(vals / self.score_temp, dim=1)

        neigh_labels = train_labels[idx]
        hard = F.one_hot(neigh_labels, num_classes=num_classes).float()
        hard_prob = torch.sum(w.unsqueeze(-1) * hard, dim=1)

        prop_prob = torch.sum(w.unsqueeze(-1) * train_prop[idx], dim=1)
        prop_prob = prop_prob / prop_prob.sum(dim=1, keepdim=True).clamp_min(1e-8)

        sim_proto = feat @ prototypes.T
        proto_prob = torch.softmax(sim_proto / self.proto_temp, dim=1)

        class_vals = torch.arange(num_classes, dtype=feat.dtype).unsqueeze(0)
        exp_severity = torch.sum(hard_prob * class_vals, dim=1, keepdim=True)
        ord_bias = -torch.abs(class_vals - exp_severity)

        fused = (
            self.fuse_w_knn_hard * torch.log(hard_prob.clamp_min(1e-8))
            + self.fuse_w_knn_prop * torch.log(prop_prob.clamp_min(1e-8))
            + self.fuse_w_proto * torch.log(proto_prob.clamp_min(1e-8))
            + self.fuse_w_ord * ord_bias
        )
        return fused

    def _build_conformal_residuals(self):
        n = self.train_feat.shape[0]
        residuals = []
        correct = 0
        for i in range(n):
            logits_i = self._fused_local_logits(
                self.train_feat[i : i + 1], leave_one_out_index=i
            )
            probs_i = torch.softmax(logits_i, dim=-1).squeeze(0)
            true_y = int(self.train_labels[i].item())
            true_prob = probs_i[true_y].item()
            residuals.append(1.0 - true_prob)
            if int(torch.argmax(probs_i).item()) == true_y:
                correct += 1

        residuals = torch.tensor(residuals, dtype=torch.float32)
        q = torch.quantile(residuals, self.conformal_q).item()
        self.residual_quantiles = torch.full(
            (self.num_classes,), float(q), dtype=torch.float32
        )
        print(
            f"local_loo_acc={correct / max(n, 1):.4f} "
            f"conformal_q={self.conformal_q:.2f} "
            f"resid_quantile={q:.4f}",
            flush=True,
        )

    def fit(self, img_dir, train_ann_file, unlabel_ann_file):
        cat_id_to_idx, idx_to_cat_id, num_classes = _coco_label_mapping(train_ann_file)
        self.idx_to_cat_id = idx_to_cat_id
        self.num_classes = num_classes

        labeled = _labeled_records(img_dir, train_ann_file, cat_id_to_idx)
        unlabeled = _unlabeled_records(img_dir, unlabel_ann_file)

        if len(labeled) == 0:
            raise RuntimeError("No labeled samples found.")
        if len(unlabeled) == 0:
            raise RuntimeError("No unlabeled samples found.")

        labeled_paths = [p for p, _ in labeled]
        yl = torch.tensor([y for _, y in labeled], dtype=torch.long)

        print(
            f"extract_features labeled={len(labeled_paths)} unlabeled={len(unlabeled)}",
            flush=True,
        )
        xl = self._extract_features(labeled_paths, batch_size=64)
        xu = self._extract_features(unlabeled, batch_size=64)

        xl = F.normalize(xl.float(), dim=1)
        xu = F.normalize(xu.float(), dim=1)

        print("running_label_propagation", flush=True)
        _, yu = self._label_propagation(xl, yl, xu)
        prop_l = self._compute_train_propagated_probs(xl, yl, xu, yu)

        self.train_feat = xl.cpu()
        self.train_labels = yl.cpu()
        self.train_prop = prop_l.cpu()
        self.class_prototypes = self._build_prototypes(
            self.train_feat, self.train_labels
        ).cpu()

        print("building_local_conformal_calibration", flush=True)
        self._build_conformal_residuals()

        with torch.no_grad():
            train_logits = self._fused_local_logits(self.train_feat)
            train_probs = torch.softmax(train_logits, dim=-1)
            train_pred = train_probs.argmax(dim=-1)
            acc = (train_pred == self.train_labels).float().mean().item()
            nll = F.nll_loss(
                torch.log(train_probs.clamp_min(1e-8)), self.train_labels
            ).item()
            print(f"fit_train_acc={acc:.4f} fit_train_nll={nll:.4f}", flush=True)

        self.model.eval()
        print("fit_done", flush=True)

    def _tta_views(self, x: torch.Tensor):
        views = [x]
        views.append(torch.flip(x, dims=[3]))
        s = min(self.tta_shift, x.shape[-1] // 16)
        if s > 0:
            views.append(torch.roll(x, shifts=-s, dims=3))
            views.append(torch.roll(x, shifts=s, dims=3))
        return views

    @torch.no_grad()
    def _predict_from_feat(self, feat: torch.Tensor):
        raw_logits = self._fused_local_logits(feat).to(self.device)
        probs = torch.softmax(raw_logits, dim=-1)

        penalty = self.residual_quantiles.to(self.device).unsqueeze(0)
        calibrated_probs = (probs - penalty).clamp_min(1e-8)
        calibrated_probs = calibrated_probs / calibrated_probs.sum(
            dim=1, keepdim=True
        ).clamp_min(1e-8)
        calibrated_logits = torch.log(calibrated_probs.clamp_min(1e-8))
        return calibrated_logits

    @torch.no_grad()
    def predict(self, image_batch):
        self.model.eval()

        if (
            self.train_feat is None
            or self.train_prop is None
            or self.residual_quantiles is None
        ):
            raise RuntimeError("Call fit() before predict().")

        if isinstance(image_batch, torch.Tensor):
            x = image_batch
            if x.ndim == 3:
                x = x.unsqueeze(0)
            x = x.float().to(self.device, non_blocking=True)
        elif isinstance(image_batch, list):
            x = torch.stack(
                [self.transform(im.convert("RGB")) for im in image_batch], dim=0
            ).to(self.device, non_blocking=True)
        else:
            raise TypeError("image_batch must be a Tensor or list of PIL images")

        views = self._tta_views(x)
        logits_sum = None
        feat_main = None

        for i, xv in enumerate(views):
            out = self.model(xv)
            feat = F.normalize(out["feat"].float(), dim=1)
            if i == 0:
                feat_main = feat
            logits_v = self._predict_from_feat(feat)
            logits_sum = logits_v if logits_sum is None else (logits_sum + logits_v)

        logits = logits_sum / float(len(views))
        pred_idx = logits.argmax(dim=-1).detach().cpu().tolist()
        predictions = [{"category_id": int(self.idx_to_cat_id[i])} for i in pred_idx]
        return logits, feat_main, predictions

"""
\end{lstlisting}
\label{fig:retino_program}
\end{promptbox}